\documentclass{article}

     \usepackage[preprint]{neurips_data_2022}
\usepackage[utf8]{inputenc} % allow utf-8 input
\usepackage[T1]{fontenc}    % use 8-bit T1 fonts
\usepackage[pagebackref]{hyperref}       % hyperlinks
\usepackage{url}            % simple URL typesetting
\usepackage{booktabs}       % professional-quality tables
\usepackage{amsfonts}       % blackboard math symbols
\usepackage{nicefrac}       % compact symbols for 1/2, etc.
\usepackage{microtype}      % microtypography
\usepackage{xcolor}         % colors
\usepackage{latexsym}
\usepackage{comment}
\usepackage{amsmath,amssymb}

\usepackage{todonotes}
\usepackage{booktabs}
\usepackage{footnote}
\usepackage{microtype}
\usepackage{graphicx,paralist,multirow,makecell}
\usepackage{booktabs}
\usepackage{color}
\usepackage{multicol}
\usepackage{subcaption}
\usepackage{xcolor,colortbl}
\usepackage{wrapfig}
\usepackage[export]{adjustbox}
\usepackage{tablefootnote}
\usepackage{diagbox}
\usepackage{slashbox}

\newcommand{\climb}{CLiMB}
\newcommand{\climbfull}{Continual Learning in Multimodality Benchmark}
\newcommand{\vilt}{ViLT}
\newcommand{\viltbert}{VAuLT}

\newcommand{\transfer}[2]{#1\% [#2]}
\definecolor{Gray}{gray}{0.9}
\newcolumntype{g}{>{\columncolor{Gray}}r}
\DeclareSymbolFont{extraup}{U}{zavm}{m}{n}
\DeclareMathSymbol{\varheart}{\mathalpha}{extraup}{86}
\DeclareMathSymbol{\vardiamond}{\mathalpha}{extraup}{87}
\newcommand{\model}{\mathcal{M}}
\newcommand{\algorithm}{\mathcal{A}}
\newcommand{\task}{\mathcal{T}}
\newcommand{\knowledgetransfer}[1]{\mathbb{T}_{\textit{UK}}(#1)}
\newcommand{\forgetting}[2]{\mathbb{T}_{F}(#1 \leftarrow #2)}
\newcommand{\lowshot}[2]{\mathbb{T}^{#1}_{LS}(#2)}
\title{\climb: A Continual Learning Benchmark\\for Vision-and-Language Tasks}

\author{
  Tejas Srinivasan$^1$ \qquad Ting-Yun Chang$^1$ \qquad Leticia Pinto Alva$^1$ \\ \textbf{Georgios Chochlakis$^1$} \qquad \textbf{Mohammad Rostami$^{1,2}$} \qquad \textbf{Jesse Thomason$^1$} \\
  $^1$University of Southern California \qquad $^2$USC Information Sciences Institute \\
  \texttt{\{tejas.srinivasan,tingyun,pintoalv,chochlak,rostamim,jessetho\}@usc.edu} 
}

\begin{document}

\maketitle

\begin{abstract}

Current state-of-the-art vision-and-language models are evaluated on tasks either individually or in a multi-task setting, overlooking the challenges of continually learning (CL) tasks as they arrive.
Existing CL benchmarks have facilitated research on task adaptation and mitigating ``catastrophic forgetting'', but are limited to vision-only and language-only tasks.
We present \climb, a benchmark to study the challenge of learning multimodal tasks in a CL setting, and to systematically evaluate how upstream continual learning can rapidly generalize to new multimodal and unimodal tasks.
\climb\ includes implementations of several CL algorithms and a modified Vision-Language Transformer (\vilt) model that can be deployed on both multimodal and unimodal tasks.
We find that common CL methods can help mitigate forgetting during multimodal task learning, but do not enable cross-task knowledge transfer.
We envision that \climb\ will facilitate research on a new class of CL algorithms for this challenging multimodal setting.
\end{abstract}

\section{Introduction}
\label{sec:intro}
Large-scale pre-trained models, including crossmodal vision-and-language models, are generally fine-tuned on each downstream task individually, requiring fine-tuning and storing new models for each task.
By contrast, multi-task learning requires fixing a set of tasks, but such training is incapable of dynamically learning new tasks. 
Although continual learning (CL) algorithms have explored cross-task knowledge transfer, existing methods primarily consider unimodal tasks in artificial settings~\citep{jin2021learn, lin2021clear}.
Multimodal pre-training can encode useful and transferable features for diverse tasks, but learning from a \emph{sequence} of different multimodal tasks and the subsequent forgetting effects~\citep{kirkpatrick2017overcoming} have not yet been studied.

Additionally, it is assumed that these deployed models will encounter tasks containing all modalities seen during training time.
This assumption means learning separate models for language-only, vision-only, and vision-language tasks, as opposed to a single ``generalist'' model that can handle all modalities or subsets of them~\citep{reed2022generalist}.
Yet, existing work suggests that knowledge grounded in multiple modalities can benefit unimodal tasks~\citep{desai2021virtex, jin2022leveraging}.
Currently, the research community lacks a suitable benchmark to systematically study how models can continually learn vision-and-language tasks while being transferable to unimodal tasks.

In this paper, we introduce the \textbf{\climbfull\ (\climb)},\footnote{The code for our benchmark is available at \url{https://github.com/GLAMOR-USC/CLiMB}} to facilitate the study of CL in vision-and-language tasks with deployment to multi- and unimodal tasks.
We formulate a learning problem wherein a model is first trained on sequentially arriving vision-and-language tasks, referred to as \textbf{upstream continual learning}, and then \textbf{transferred downstream to low-shot} multimodal and unimodal tasks.
\climb\ initially includes four vision-and-language tasks, five language tasks, and four vision tasks, and is extensible to new tasks, models, and learning algorithms.
Experiments using \climb\ find that existing CL algorithms can mitigate forgetting, but not transfer knowledge across tasks, revealing a need for new research into continual learning strategies for vision-language tasks.
Further, current CL algorithms and multimodal models are not well suited for low-shot adaptation to multimodal or unimodal tasks.
We hope \climb\ will provide the basis for developing models and learning algorithms for multimodal continual learning.

\section{Background and Related Work}
\label{sec:rel-work}
\begin{figure}[t]
    \centering
    \includegraphics[width=.9\columnwidth]{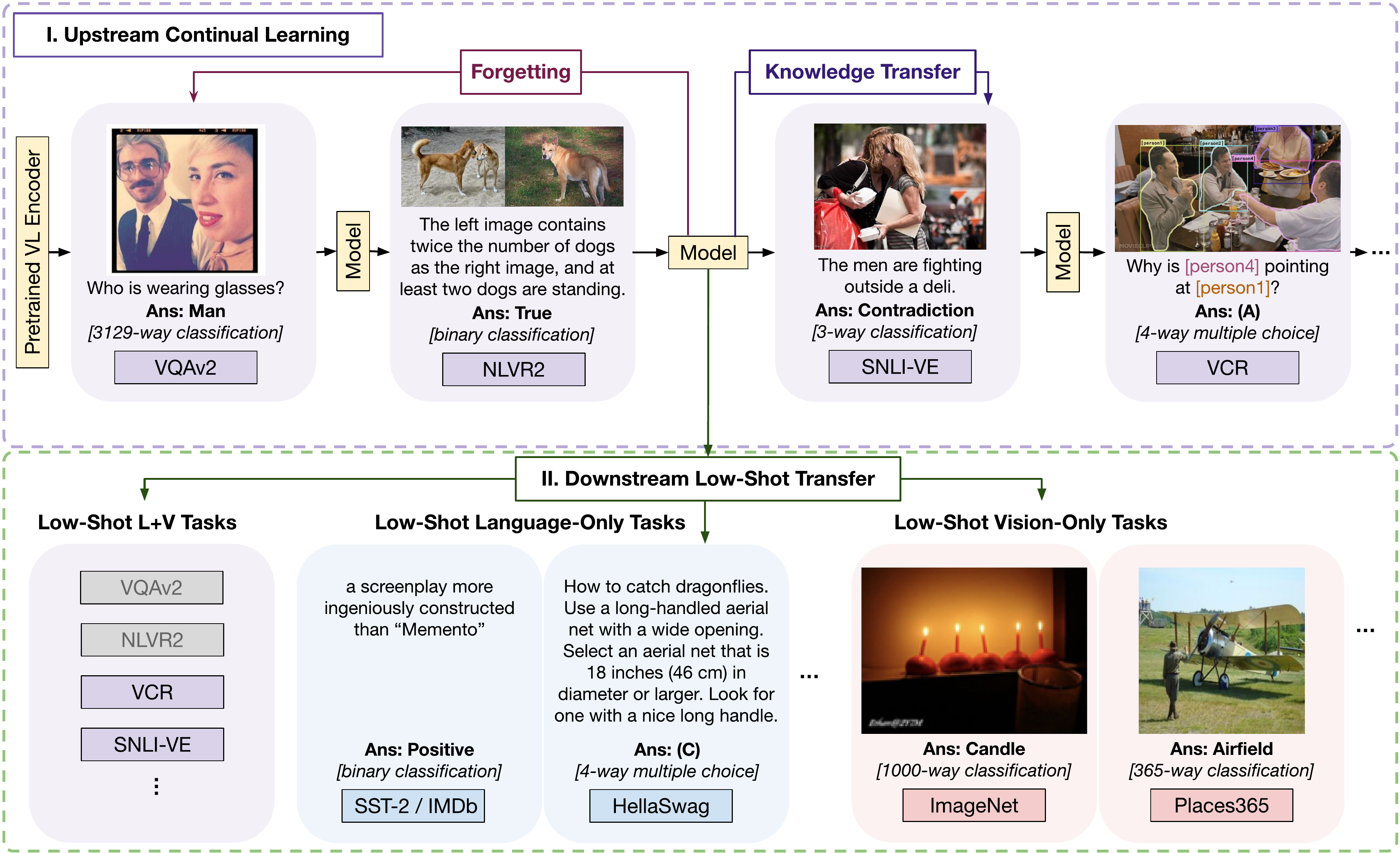}
    \caption{\climb\ evaluates candidate CL models and learning algorithms in two phases.
    For Phase I, Upstream Continual Learning, a pre-trained multimodal model is trained on a sequence of vision-and-language tasks, and evaluated after each task on its degree of Forgetting of past tasks and Knowledge Transfer to the next task. 
    For Phase II, after each multimodal task the model is evaluated for its Downstream Low-Shot Transfer capability on both multimodal and unimodal tasks.
    }
    \label{fig:cl-pipeline}
\end{figure}

Continual, or lifelong, learning is an essential ability to develop autonomous agents that can learn in a cumulative way~\citep{chen2018lifelong}.
In CL, a model is trained on sequentially arriving tasks and evaluated both on its ability to learn future tasks as well as retain performance on past learned tasks~\citep{kirkpatrick2017overcoming}. 
A necessity for developing CL algorithms is benchmarks that collate suitable sequential tasks.
There are two primary approaches to create such CL benchmarks.

The first approach is to split existing tasks into non-overlapping sub-tasks that are sequentially presented for continual learning.
For example, one can divide tasks along input categories~\citep{greco2019psycholinguistics} or output classes~\citep{vinyals2016matching,kirkpatrick2017overcoming} into disjoint sets.
Mimicking real world distribution shift, timestamps can group data instances according to the order of their creation~\citep{lin2021clear}.

\climb\ goes beyond such artificial, single-task-based CL and instead aggregates several diverse tasks.
Similarly, unimodal benchmarks such as Visual Domain Decathlon~\citep{rebuffi2017learning} and Natural Language Decathlon~\citep{mccann2018natural} aggregate 10 image classification and 10 language tasks, respectively.
The CLIF-26 benchmark~\citep{jin2021learn} is built for CL on the GLUE~\citep{wang2018glue} language tasks.
\climb\ goes beyond these unimodal benchmarks by evaluating on sequences of multimodal, vision-and-language tasks \textit{and} testing downstream transfer to unimodal tasks.

\clearpage

\section{\climb: The Continual Learning in Multimodality Benchmark}
\label{sec:cl}
\begin{table}[t]
  \centering
  %\begin{scriptsize}
  \small
  \begin{tabular}{lllll}
    \toprule
    \bf Task & \bf Vision Input(s) & \bf Language Input(s) & \bf Decision & \bf Score Metric \\
    \midrule
    VQAv2 & Image & Question & 1 of 3129 & VQAScore\tablefootnote{https://visualqa.org/evaluation.html} \\
    NLVR2 & 2 images & Caption & True/False & Accuracy \\
    SNLI-VE & Image & Hypothesis & Ent/Neu/Con & Accuracy \\
    VCR & Image w/ bboxes & Question + 4 Answers & 1 of 4 & Accuracy\\
    \midrule
    IMDb & & Sentence & Pos/Neg & Accuracy\\
    SST-2 & & Sentence & Pos/Neg & Accuracy\\
    HellaSwag & & Sentence Prefix + 4 Endings & 1 of 4 & Accuracy \\
    CommonsenseQA & & Question + 5 Answers & 1 of 5 & Accuracy \\
    PIQA & & Question + 2 Answers & 1 of 2 & Accuracy \\
    \midrule
    ImageNet-1000 & Image & & 1 of 1000 & Top-1 Accuracy \\
    iNaturalist2019 & Image & & 1 of 1010 & Top-1 Accuracy \\
    Places365 & Image & & 1 of 365  & Top-1 Accuracy\\
    COCO-object & Image & & $n$ of 80 & Micro-F1\\
    \bottomrule
  \end{tabular}
  %\end{scriptsize}
  \small
  \caption{
    The initial tasks in \climb\ include various forms of vision and language inputs, and each task is framed as a classification problem.
    Multimodal vision-and-language tasks serve as upstream training for both multimodal and unimodal downstream, low-shot tasks.
  }
  \label{tab:benchmark}
\end{table}

\climb\ tests the ability of models and learning algorithms to adapt to a sequentially arriving stream of vision-language tasks, as well as rapidly transfer to new multimodal and unimodal tasks (Table~\ref{tab:benchmark}).

\subsection{\climb\ Learning and Evaluation}

Learning and evaluation in \climb\ proceeds in two phases: \textbf{upstream continual learning} and \textbf{downstream low-shot transfer} (Figure~\ref{fig:cl-pipeline}).
Table~\ref{tab:eval} summarizes our CL evaluation metrics.
We denote a task with modality $M \in \{V, L, VL\}$ as $\task_{M}^i$ and the number of such tasks as $K_M$.

\paragraph{Upstream Continual Learning of Multimodal Tasks}
A candidate model $\model$ encounters a sequence of $K_{VL}$ vision-language tasks, $\task_{VL}^{1\dots K_{VL}}$. 
$\model$ can be initialized with a pre-trained encoder.
We allow parameter additions to the base model on a per-task basis.
In this work we add task-specific classification layers for each new task on top of the base encoder model.
The model $\model$ is sequentially trained on the training split of each task $\task_{VL}^i$ with candidate CL algorithm $\algorithm$.
For task $\task_{VL}^i$, the model is not presented with any inputs from $\task_{VL}^{1\dots i-1}$.
However, the CL algorithm $\algorithm$ may allocate memory to access previous training examples.

We evaluate two primary model properties in the upstream phase: \textbf{upstream knowledge transfer} from past learned tasks to new tasks, and withstanding \textbf{forgetting} of previously-seen tasks.
The upstream knowledge transfer $\knowledgetransfer{i}$ on task $\task_{VL}^i$ is the relative gain in score from learning the previous tasks $\task_{VL}^{1...i-1}$.
Forgetting $\forgetting{j}{i}$ of previously-seen task $\task_{VL}^j$ is the relative performance degradation in that task after learning subsequent tasks $\task_{VL}^{j+1...i}$ (Table~\ref{tab:eval}). 

\paragraph{Downstream Transfer to Low-Shot Tasks}
We evaluate the low-shot adaptation ability of the model $\model$ after learning  each upstream vision-language task.
After training on the $i^{th}$ upstream task $\task_{VL}^i$, we evaluate low-shot transfer to remaining multimodal tasks $\task_{VL}^{i+1...K_{VL}}$, as well as unimodal tasks $\task_{V}^{1...K_V}$ and $\task_{L}^{1...K_L}$.
Specifically, for every task in each modality, a low-shot instance of task $\task_M^i$, denoted as $\task_{M}^{LS(i)}$, is created.
The \textbf{low-shot transfer} ability to this task is evaluated by fine-tuning upstream encoder checkpoints on task $\task_{M}^{LS(i)}$. 
We compute the low-shot transfer $\lowshot{M}{i}$ as the relative improvement of the CL encoder's performance on the low-shot task $\task_{M}^{LS(i)}$, denoted as $S_\algorithm^{LS(i)}$, over the pre-trained encoder's performance on the same low-shot task, $S_{PT}^{LS(i)}$.

\begin{table}[t]
    \setlength{\aboverulesep}{0pt}
    \setlength{\belowrulesep}{0pt}
    \centering
    \small
    \begin{tabular}{>{\raggedright\arraybackslash}p{0.17\linewidth}p{0.49\linewidth}p{0.24\linewidth}}
     Evaluation Type
     & Description & Metric ($\times 100\%$) \\
    \midrule
        Upstream Knowledge Transfer, $\knowledgetransfer{i}$ & Improvement of performance on task $\task_{VL}^i$ after training on tasks $\task_{VL}^{1...i}$ using algorithm $\algorithm$ ($S_\algorithm^i$) compared to finetuning the pretrained model on $\task_{VL}^i$ directly ($S_{PT}^i$). & $\knowledgetransfer{i}  = \frac{S_{\algorithm}^i - S_{PT}^i}{S_{PT}^i - S_{R}^i}$ \\
    \midrule
        Forgetting Transfer, $\forgetting{j}{i}$ &
        Decrease of performance when a model trained on tasks $\task_{VL}^{1...i}$ is evaluated on task $\task_{VL}^j (j < i)$. $S_{\algorithm}^{j \leftarrow i}$ denotes model performance on $\task_{VL}^j$ after training up to $i$.
        & $\forgetting{j}{i} = \frac{S_{\algorithm}^j - S_{\algorithm}^{j \leftarrow i}}{S_{\algorithm}^j - S_{R}^j}$\\
    \midrule
        Low-Shot Transfer, $\lowshot{M}{i}$ &
        Improvement of performance on low-shot task $\task_{M}^{LS(i)}$ using an encoder checkpoint trained by upstream algorithm $\algorithm$ ($S_\algorithm^{LS(i)}$) compared to learning the same low-shot task without any upstream learning ($S_{PT}^{LS(i)}$). & 
        $\lowshot{M}{i} = \frac{S_{\algorithm}^{LS(i)} - S_{PT}^{LS(i)}}{S_{PT}^{LS(i)} - S_{R}^i}$\\
    \bottomrule
    \end{tabular}
    %\end{scriptsize}
    \small
    \caption{
        Model and algorithm evaluation metrics in the upstream and downstream phases.
        For the $i^{th}$ task, we compute a model's $\delta^i=S^i-S_R^i$, the model's task score $S^i$ minus the score $S_R^i$ of random selection. 
        Our evaluation protocol computes each metric as a relative change in the $\delta^i$ of a CL algorithm $\algorithm$ over a baseline setting $\mathcal{B}$ to enable across-task comparisons. 
        Each evaluation metric is calculated as $\frac{\delta_\algorithm - \delta_\mathcal{B}}{\delta_\mathcal{B}} = \frac{S_\algorithm - S_\mathcal{B}}{S_\mathcal{B} - S_R}$, and is presented as a percentage.
    }
\label{tab:eval}
\end{table}

\subsection{\climb\ Multimodal and Unimodal Tasks}

\climb\ initially includes four multimodal upstream vision-language tasks, five language-only tasks, and four vision-only tasks (Table~\ref{tab:benchmark}).
We frame each as a classification task.

\paragraph{Vision-Language Tasks}
\climb\ includes VQAv2~\citep{goyal2017making}, NLVR2~\citep{suhr2018corpus}, SNLI-VE~\citep{xie2019visual} and VCR~\citep{zellers2019vcr}. 
Solving these challenging tasks requires different kinds of knowledge in the multimodal model: question answering, visual and commonsense reasoning, entailment understanding.

\paragraph{Language-Only Tasks}
\climb\ includes IMDb~\citep{maas-EtAl:2011:ACL-HLT2011}, SST-2~\citep{socher2013recursive}, HellaSwag~\citep{zellers2019hellaswag}, CommonsenseQA~\citep{talmor2018commonsenseqa}, and PIQA~\citep{bisk2020piqa}.
We hypothesize that visually-grounded knowledge from upstream tasks may benefit tasks such as IMDb and SST-2, which are sourced from movie reviews, as well as HellaSwag, sourced from video captions, and PIQA, sourced from physically-grounded instructions from images and videos.
Further, commonsense knowledge and reasoning skills obtained from VCR and NLVR2 may benefit tasks like HellaSwag, CommonsenseQA, and PIQA.

\paragraph{Vision-Only Tasks}
\climb\ includes ImageNet-1000~\citep{russakovsky2015imagenet}, iNaturalist2019~\citep{van2018inaturalist}, Places365~\citep{mahajan2018exploring}, and MS-COCO object detection (formulated as a multi-label classification task). 
%  similar to prior work~\citep{he2021masked}.
Since VQAv2 images are sourced from MS-COCO~\citep{lin2014microsoft}, we hypothesize VQAv2 upstream learning may aid in the COCO object detection task.

\section{Modeling and Experiments}
\label{sec:experiments}
Using \climb, we study the performance of several commonly used CL algorithms on multimodal tasks.
We use fixed upstream task order (VQAv2 $\rightarrow$ NLVR2 $\rightarrow$ SNLI-VE $\rightarrow$ VCR).

\subsection{Vision-Language Encoder: ViLT}
\label{subsec:vilt}
\begin{figure}[t]
    \centering
    \includegraphics[width=0.9\columnwidth]{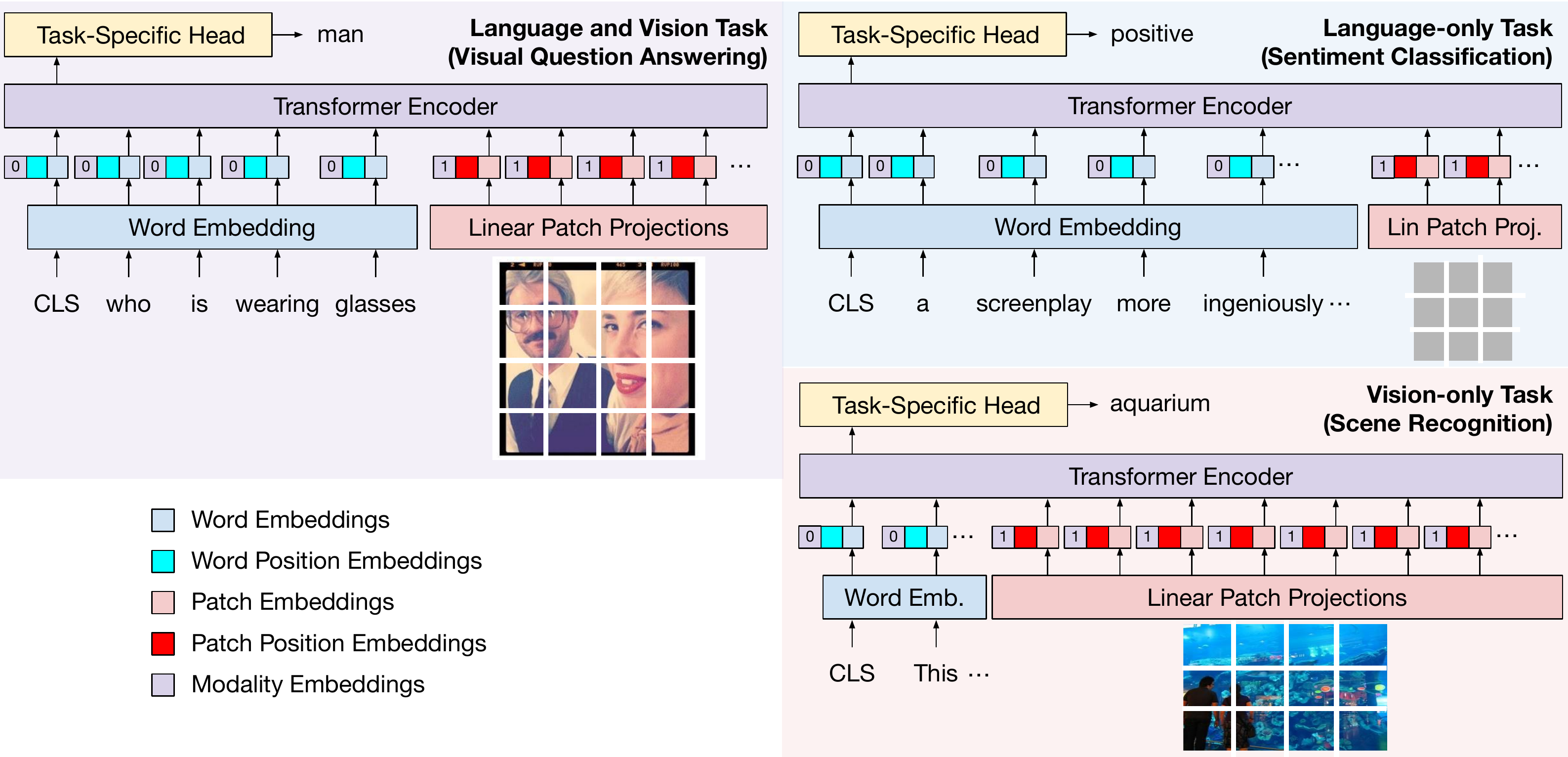}
    \caption{
        The \vilt\ model~\citep{kim2021vilt} operates on vision and language inputs (left). 
        We adapt these inputs for language-only tasks by providing the average MS-COCO image as in-domain visual input, and vision-only tasks by providing vacuous language input ``This is an image'' (right).
    }
    \label{fig:vilt}
\end{figure}

We use a pre-trained Vision-Language Transformer (\vilt)~\citep{kim2021vilt} as a backbone encoder.
Unlike other pre-trained vision-language models~\citep{lu2019vilbert,chen2019uniter} that build upon region-level features extracted from Faster R-CNN~\citep{ren2015faster}, \vilt\ directly operates on image patches without using any convolutional layers. 
In \vilt, text tokens and image patches are concatenated into an input sequence and passed through a Transformer, which learns the vision-language alignment with self-attention across both modalities (Figure~\ref{fig:vilt}). 

\subsection{Upstream Experiments: Algorithms and Task Ordering}
\label{subsec:algs}

\climb\ includes several CL algorithm implementations.
\textbf{Sequential Fine-tuning (SeqFT)} fine-tunes the full encoder and task-specific layers for each task in order. 
This baseline algorithm is an extension of the single-task fine-tuning paradigm to the CL setting.
We also experiment with a \textbf{Frozen Encoder} baseline that trains only the task-specific layers. 
Fine-tuning all parameters may cause forgetting since the encoder parameters are overwritten, while fine-tuning only the task-specific layer prevents knowledge transfer since the shared encoder parameters are fixed.
In \textbf{Frozen Bottom-K}, we freeze the bottom $K$ layers of the encoder and fine-tune the rest, balancing these solutions (we set $K$=9).

\climb\ also includes two CL algorithms that fine-tune all parameters but are designed to mitigate forgetting.
\textbf{Experience Replay (ER)}~\citep{chaudhry2019tiny} caches a small percentage of training examples in a memory buffer after training on each task, and periodically performs a ``replay'' training step using cached examples to refresh the model.
\textbf{Elastic Weight Consolidation (EWC)}~\citep{kirkpatrick2017overcoming} is a regularization method that adds an auxiliary L2 loss between weights in the current model and previous checkpoints to slow change in important encoder parameters.

Finally, \climb\ includes \textbf{Adapters}~\citep{houlsby2019parameter}, which add a small number of task-specific parameters, called Adapter modules, within layers of the pre-trained encoder.
During training, the encoder's original parameters are kept frozen and only the Adapter modules are trained. 
We use a new set of Adapter modules each time we train on a new task, which leaves the previous tasks' modules untouched and prevents forgetting, but also does not facilitate cross-task knowledge transfer.

\subsection{Downstream Low-Shot Experiments}

We consider low-shot multimodal and unimodal tasks.
We first define low-shot settings for different task types, then explain how we apply the multimodal \vilt\ model to unimodal settings.

\paragraph{Low-Shot Task Settings} We study ``low-shot'' training paradigms where only a fraction of full training data is available.
For the multimodal classification tasks, NLVR2 and SNLI-VE, we use 2048 examples per class, whereas for the multiple choice VCR task, we use 5\% of the training data.
Among unimodal tasks, for vanilla classification tasks (IMDb, SST2, ImageNet-1000, iNaturalist, and Places365), we consider training with $N=\{16, 32\}$ examples per class.
For multiple choice classification tasks (PIQA, CommonsenseQA, HellaSwag), we use $N=\{1024, 4096\}$ since these tasks are considerably more challenging.
For the multi-label COCO object detection task, we consider $M=\{5\%, 10\%\}$ of the original training data.

\begin{table}[t]
    \setlength{\aboverulesep}{0pt}
    \setlength{\belowrulesep}{0pt}
    \centering
    \begin{small}
    \begin{tabular}{lgrrrrr}
    \toprule
    \multirow{2}{*}{Alg $\algorithm$} & Params &  \multicolumn{1}{c}{Task 1} & \multicolumn{1}{c}{Task 2} & \multicolumn{1}{c}{Task 3} & \multicolumn{1}{c}{Task 4} \\
    & Trained &  \multicolumn{1}{c}{VQAv2} & \multicolumn{1}{c}{NLVR2} & \multicolumn{1}{c}{SNLI-VE} & \multicolumn{1}{c}{VCR} \\
    \midrule
    Direct FT & 100\% &  [67.70]	& [73.07] & [76.31] & [61.31] \\
    \midrule
    SeqFT  & 100\% &  \transfer{0.13}{67.79}	& \transfer{-1.80}{72.66} & \transfer{-3.33}{74.89} & \transfer{-5.09}{59.47} \\
    Frozen Enc & 7.88\% &  \transfer{-14.10}{58.15} & \transfer{-40.78}{63.66} & \transfer{-15.98}{69.45} & \transfer{-53.47}{41.90} \\
    Frozen B9 & 25.92\% &  \transfer{-0.58}{67.30} & \transfer{-0.58}{72.94} & \transfer{-3.31}{74.90} & \transfer{-15.49}{55.69} \\
    ER & 100\% &  \transfer{0.26}{67.87} & \transfer{0.56}{73.20} & \transfer{-2.89}{75.08} & \transfer{-4.45}{59.70} \\
    EWC & 100\% &  \transfer{0.20}{67.84} & \transfer{-2.79}{72.39} & \transfer{-4.52}{74.38} & \transfer{-4.86}{59.55} \\
    Adapters & 13.02\% &  \textbf{\transfer{0.59}{68.10}} & \textbf{\transfer{2.55}{73.66}} & \textbf{\transfer{-0.56}{76.08}} & \textbf{\transfer{-0.36}{61.18}} \\
    \bottomrule
    \end{tabular}
    \end{small}
    \caption{
        Upstream Knowledge Transfer $\knowledgetransfer{i}$ relative to direct fine-tuning on each task, along with task score $[S_{\algorithm}^i]$ (\%), for different CL algorithms $\algorithm$ applied to \vilt.
        No CL algorithms achieve notable positive Knowledge Transfer, while the majority in fact \emph{hurt} learning of new tasks.
    }
\label{tab:knowledge-transfer}
\end{table}

\paragraph{Unimodal Tasks in \vilt}
To apply \vilt\ to vision-only tasks, we use the phrase ``This is an image'' as the language input paired with the input image from the vision task.
For language-only tasks, however, we need to address several challenges to effectively apply \vilt.

First, we find that averaging all MS-COCO training images into a single, in-distribution image paired with text inputs produces better results with \vilt\ than not concatenating any image tokens at all to the Transformer input sequence.

Second, \vilt\ only allows a maximum of 40 language tokens in the input, which is enough for captions but insufficient for most language tasks. 
To deal with this challenge, we first downsample the vacuous image to reduce its token length from $144$ to $16$. 
Next, we extend the available language tokens by creating copies of pre-trained \vilt's language positional embeddings, $E\in \mathbb{R}^{40\times d}$, and concatenating these copies to get extended positional embeddings, $\hat{E}\in \mathbb{R}^{L\times d}$, where $L$ is the maximum sequence length of each task and $d$ is the embedding dimension.

Finally, \vilt’s language pre-training is on image captions that do not represent more general language use.
We additionally experiment with a \viltbert~\citep{viltbert} model that extracts language representations from a pre-trained, frozen BERT~\citep{devlin2018bert} that serve as input embeddings for \vilt. 
Please refer to the supplementary materials for more experiments and details.

\section{Results}
\label{sec:results}

We present Knowledge Transfer and Forgetting capabilities of different CL algorithms, experiments with multiple upstream task orders, and Low-Shot Transfer to downstream tasks.

\subsection{Upstream Learning Results}
We find that common CL algorithms do not facilitate positive knowledge transfer in the vision-and-language setting of \climb, and in fact often hurt future task learning.
Some are able to effectively mitigate forgetting, but none perform as well as directly fine-tuning on a candidate task.
By examining the effects of task order, we conclude that the VCR task hurts further upstream task learning.

\begin{figure}
    \centering
    \begin{subfigure}[t]{0.4\textwidth}
      \centering
      \includegraphics[width=\linewidth]{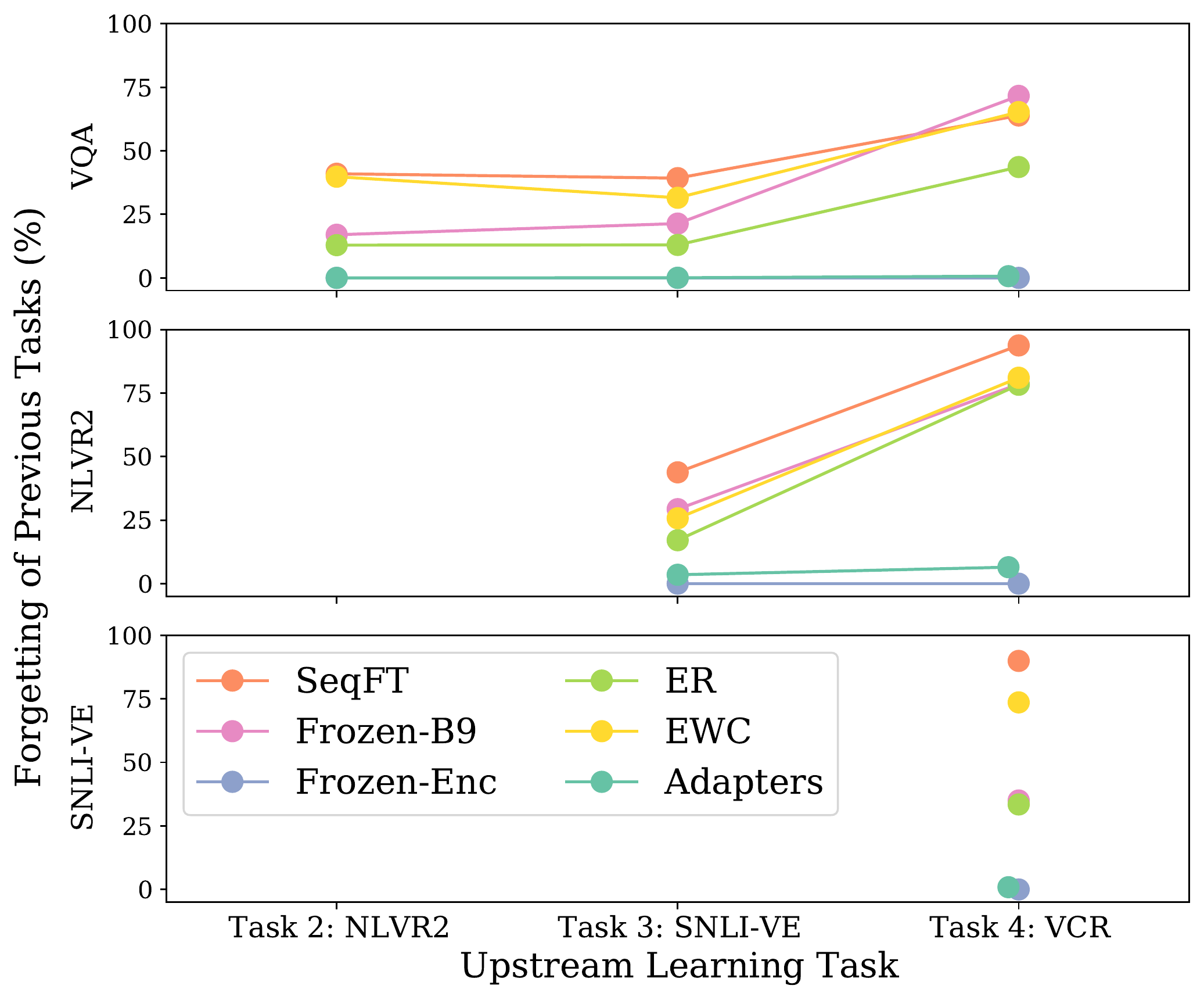}
      \caption{}
      \label{subfig:forgetting}
    \end{subfigure}
    \begin{subfigure}[t]{0.56\textwidth}
      \centering
      \includegraphics[width=\linewidth]{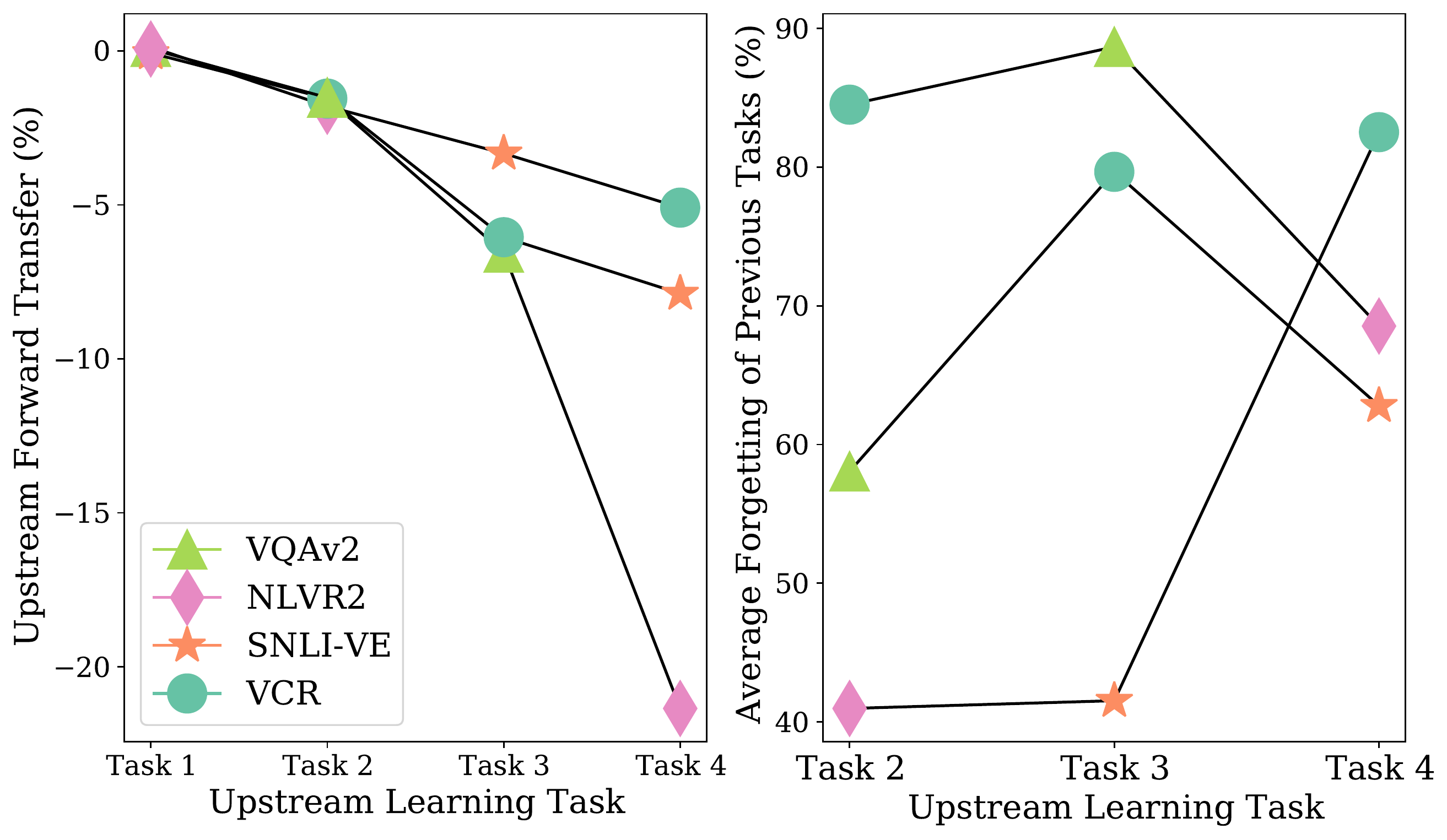}
      \caption{}
      \label{subfig:taskorder}
    \end{subfigure}
    \caption{
        \textbf{(a)} Forgetting $\forgetting{j}{i}$ (\%) of the previous $i-1$ tasks for each algorithm. 
        Each subplot denotes model performance on one of the previous tasks. 
        While all algorithms that fine-tune shared parameters exhibit Forgetting, ER best preserves past task performance.
        \textbf{(b)} Effect of task order on upstream Knowledge Transfer (left) and Forgetting (right) for three different orders.
        Lines represent performance conditioned on a particular task order.
        After experiencing the VCR task, models exhibit lower Knowledge Transfer and higher Forgetting.
    }
\end{figure}

\paragraph{Upstream Knowledge Transfer} 
In Table~\ref{tab:knowledge-transfer}, we compare the upstream knowledge transfer exhibited by the different algorithms described in Section~\ref{subsec:algs}. 
Freezing the entire encoder severely underperforms the direct fine-tuning baseline for each task.
Among other methods, all perform similarly to directly fine-tuning on the first task, with approximately zero knowledge transfer. 
However, for all methods other than Adapters, more continual learning hurts the model's ability to learn new tasks, as evidenced by the increasingly negative upstream transfer for later tasks. 
This effect may be due to loss of pre-training knowledge which is useful for task adaptation. 
This property of models to learn new tasks poorly in a continual learning setting is also called intransigence~\citep{chaudhry2018riemannian}. 
Adapters, which do not train shared encoder parameters, do not exhibit this negative knowledge transfer, and show comparable performance to full model fine-tuning despite having very few learnable parameters.

\begin{wrapfigure}[17]{r}{0.5\textwidth}
    \centering
    \includegraphics[width=\linewidth]{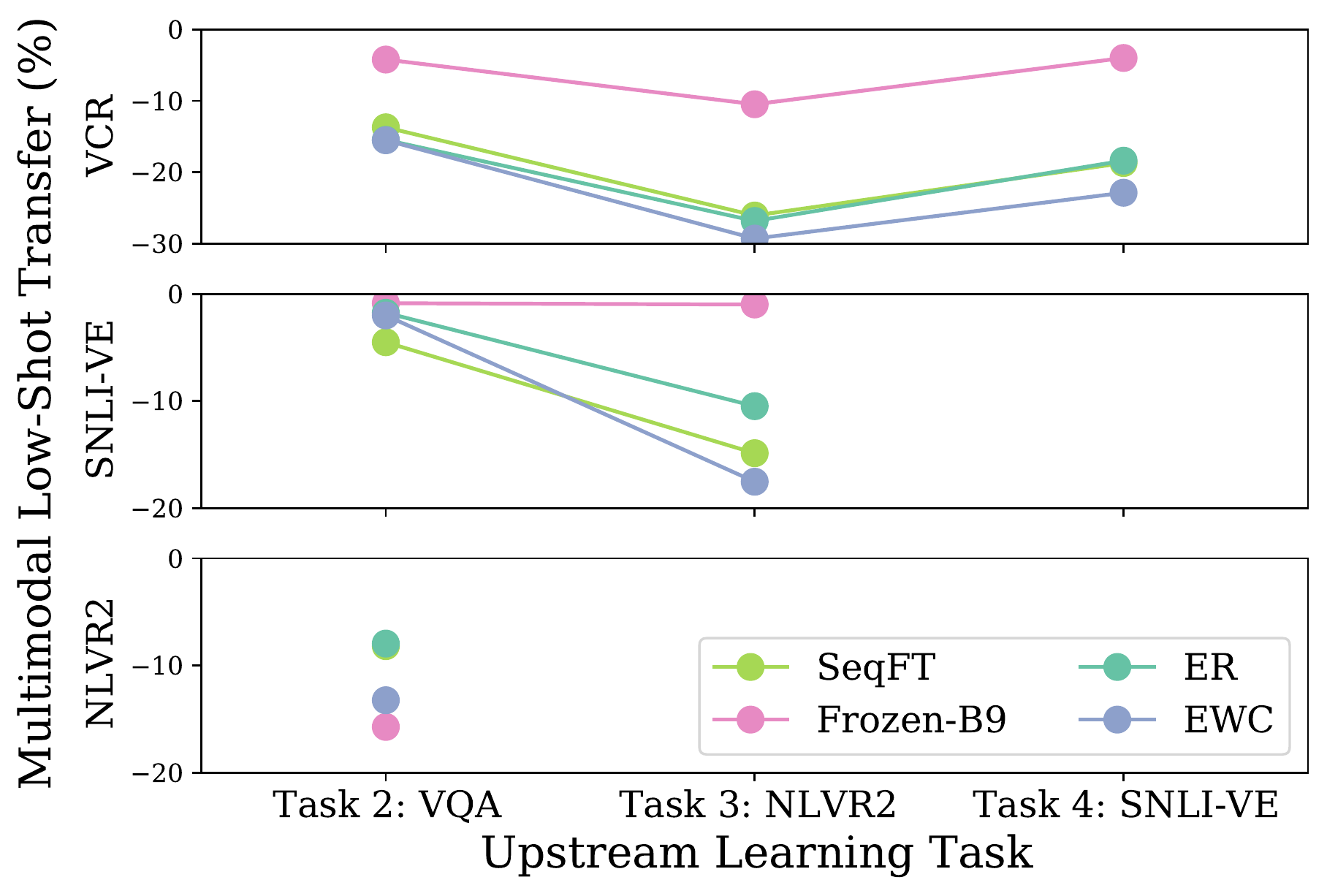}
    \caption{
        Low-shot transfer, $\lowshot{VL}{j}$, for multimodal tasks $j=\{i+1,...,K_{VL}\}$ after training on upstream task $\task_{VL}^i$.
        All CL algorithms exhibit negative Low-shot transfer on all multimodal tasks.
    }
    \label{fig:lowshot-multimodal}
\end{wrapfigure}

\paragraph{Forgetting}
Figure~\ref{subfig:forgetting} shows how each algorithm affects forgetting of previous tasks. 
Sequential Fine-tuning forgets previous tasks to a large extent, Frozen Bottom-9 shows slight improvement, and freezing the encoder prevents forgetting entirely.
Experience Replay does a better job at retaining task performance, while EWC shows only a slight improvement.
Adapters enable a model to learn upstream tasks in the multimodal CL setting \emph{while not forgetting tasks already learned}, adding only 3-4\% parameters per task.
Interestingly, forgetting is more severe after training models on the VCR task, demonstrating that the order of encountering tasks affects continual learning.

\paragraph{Effect of Upstream Task Order}
Figure~\ref{subfig:taskorder} shows the upstream knowledge transfer and forgetting for \vilt\ using Sequential Fine-tuning on three different task orders.
While the upstream transfer is similar for the first two tasks in each task ordering, training on VCR negatively affects both knowledge transfer to future tasks and forgetting of past tasks.
This effect may be due to a shift in the visual domain of VCR: input images have colored boxes drawn on them to represent grounded objects in the question, following previous work~\citep{zellers2021merlot, hessel2022abduction}.

\subsection{Downstream Low-Shot Transfer Results}

In downstream transfer, we fine-tune the entire model irrespective of the upstream CL algorithm. 
We find that upstream learning with current CL algorithms\footnote{We do not include Adapters and Frozen-Encoder as they do not modify pre-trained \vilt's parameters.} does not help the \vilt\ encoder generalize to multimodal and unimodal tasks in low-shot settings.

\begin{figure}[t]
    \centering
    \includegraphics[width=1\linewidth]{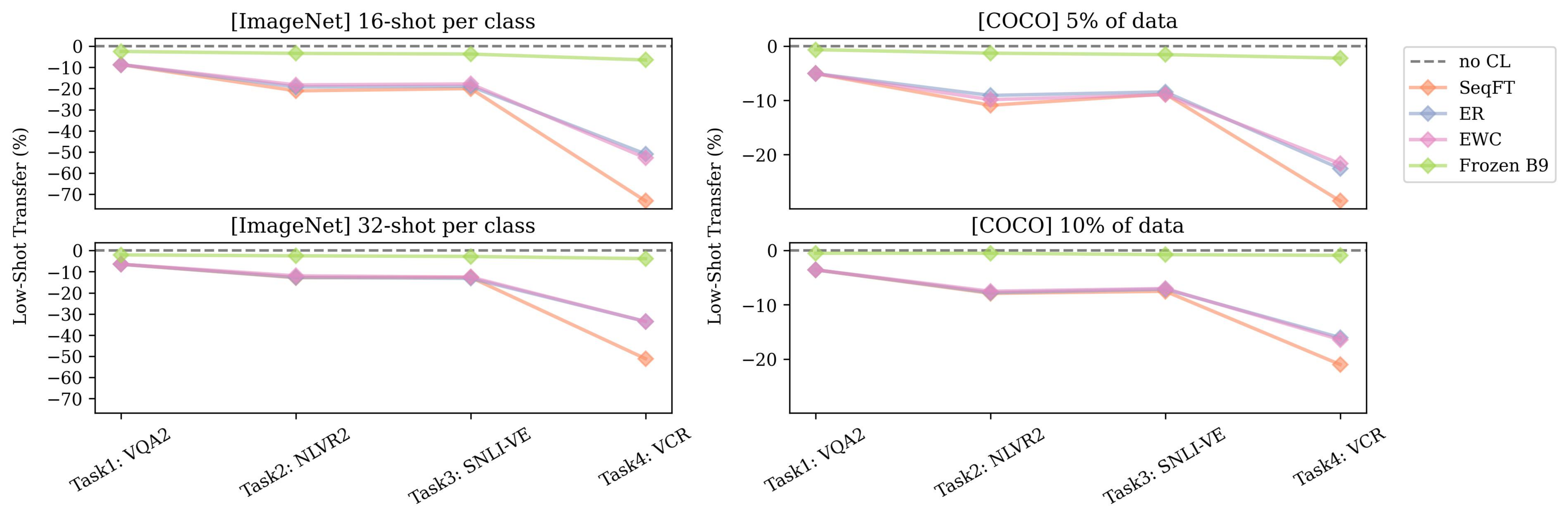}
    \caption{
        Low-Shot Transfer (\%) comparison between different CL algorithms on downstream vision-only tasks (left: ImageNet; right: COCO). 
        Findings on iNaturalist 2019 and Places365 are similar to ImageNet (see Supp).
        Generally, current CL algorithms hurt low-shot transfer compared to direct fine-tuning, with Frozen Bottom-9 being the least harmful.
    }
    \label{fig:vision-only}
\end{figure}

\paragraph{Vision-Language Tasks}
Figure~\ref{fig:lowshot-multimodal} presents the low-shot transfer $\lowshot{VL}{j}$ for all future tasks $j > i$ after training on upstream task $\task_{VL}^i$ (x-axis). 
We observe that low-shot transfer is always negative, implying that upstream continual learning always hurts the model's ability to learn new tasks in low-shot settings. 
Since upstream learning hurts model adaptation on new multimodal tasks with full training data (Table~\ref{tab:knowledge-transfer}), it is expected that this effect will also be reflected in the low-shot regime.

\paragraph{Vision-Only Tasks} 
Figure~\ref{fig:vision-only} presents low-shot transfer on vision downstream tasks, using checkpoints from different upstream CL algorithms.
Fine-tuning \vilt\ without CL performs well on vision-only tasks, achieving $65\%$ top-1 accuracy on ImageNet-1000 with only 16 shots per class (see Supp).
This performance suggests that \vilt\ already contains rich visual representations, making it sample efficient when transferred to vision-only tasks.

Second, CL actually \emph{hurts} the transferability to downstream vision tasks.
Among CL algorithms, Sequential Fine-tuning is the most harmful one, while freezing the bottom 9 layers causes the least degradation, almost matching direct fine tuning.
This finding is consistent with previous work suggesting that bottom layers in deep models learn more general and transferable representations than upper layers~\citep{yosinski2014transferable,lee2019would}.

Notably, upstream VQA and SNLI-VE checkpoints have a less negative effect on downstream COCO performance compared to NLVR2 and VCR.
Because images from NLVR2 and VCR are more dissimilar to MS-COCO than the image sources of VQA and SNLI-VE, we hypothesize that large data distribution shifts between upstream and downstream tasks hurts transfer.

\begin{figure}[t]
    \centering
    \includegraphics[width=1\linewidth]{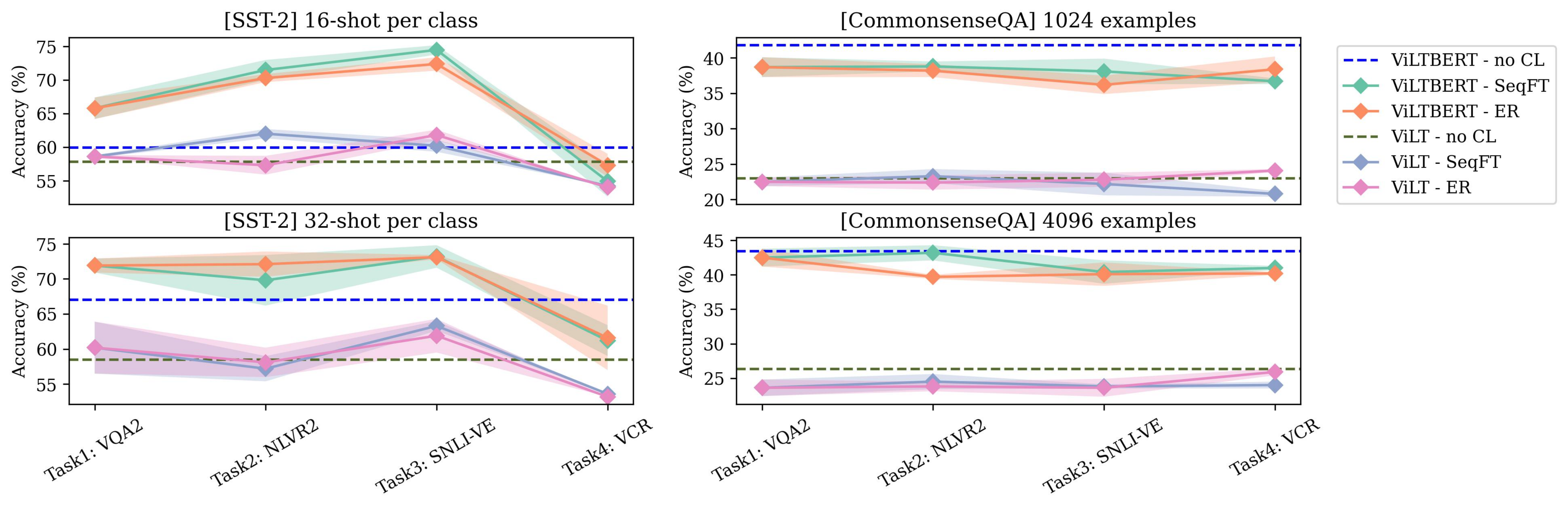}
    \caption{
        Comparisons between different encoders and continual learning algorithms on downstream language-only tasks (left: SST-2; right: CommonsenseQA).
        Note that the proposed Low-Shot Transfer metric is computed relative to the pre-trained encoder, making scores for different encoders (in this case, \vilt\ and \viltbert) incomparable. 
        Hence, we plot the absolute accuracy with shaded standard deviation.
        \viltbert\ strictly improves absolute accuracy over \vilt\ in direct fine-tuning and CL settings.
        See Supp for comparable IMDb, HellaSwag, and PIQA results.
    }
    \label{fig:lang-only}
\end{figure}

\paragraph{Language-Only Tasks}
In Figure~\ref{fig:lang-only}, we compare the performance of two pre-trained encoders, \vilt\ and \viltbert, on low-shot language tasks, and the effects of upstream multimodal CL on low-shot transfer when applied to both encoders. 

We observe that model adaptation to language tasks is challenging.
The \vilt\ model frequently performs only marginally better than the random baseline, regardless of the upstream algorithm. 
Using \viltbert\ as the encoder achieves notably higher accuracy compared to \vilt\ on all tasks, indicating that strong language priors are key to low-shot language adaptation.

All upstream CL tasks improve \viltbert's transferability to SST-2 except for VCR.
For both SST-2 and IMDb, there are significant drops after learning VCR in the upstream phase with \vilt\ and \viltbert, following vision-only results showing VCR is farther out of distribution than other upstream tasks.

However, we do not observe similar trends on the three multiple-choice tasks, where CL generally hurts.
We believe that current multimodal tasks do not learn complex language reasoning skills, hurting model transferability to language-only reasoning tasks.

\section{Conclusions}
\label{sec:conclusions}
We propose the \textbf{\climbfull\ (\climb)} to study CL in multimodal tasks with deployment to multi- and unimodal tasks.
Our experiments find that existing CL strategies do not generalize well to sequences of multimodal tasks or enable effective low-shot adaptation to downstream multi- or unimodal tasks.
We hope \climb\ will allow systematic evaluation of new models and algorithms for multimodal continual learning.
\climb\ is designed to be an extensible community tool for studying tasks, model architectures, and CL algorithms.

%Further, we currently only perform experiments with a \vilt-based model, which directly operates on image patches, unlike other vision-language encoders which use pre-extracted object features~\citep{chen2019uniter,lu2019vilbert}. 
%However, our benchmark is model-agnostic and can be extended to models using object features.

\section{Limitations}
\label{sec:limitations}
\paragraph{Task-Specificity} The current \climb\ design allows for task-specific parameters and for model awareness of the task, but multi-task language modelling has seen impressive results from reframing all tasks as sequence-to-sequence problems that remove task-specific parameters~\citep{2020t5}.
In future iterations of \climb, we intend to explore this task-agnostic paradigm, building further on the promising Adapters methods by learning a library of adapters that are dynamically selected based on input vision and language on a per-instance basis.
Additionally, the task formulations in \climb\ are mostly classification, but sequence-based vision-and-language tasks could allow the study of embodied navigation~\citep{anderson2018vision} and task completion~\citep{shridhar:cvpr20}, and may be feasible in a more task-agnostic \climb\ framework.

\paragraph{Additional CL Metrics} We have defined a set of metrics and methodologies for the challenge of multimodal continual learning, but these metrics are only an initial starting point. 
We design CLiMB to be flexible so that researchers can add metrics that they find valuable to measure, such as intransigence~\citep{chaudhry2018riemannian}.

\paragraph{Ethical Considerations} The initial \climb\ release is limited to English-only text, eliding the challenges of multi-lingual language tasks.
Further, images in currently included datasets are sourced from social media, movie clips and web searches, thus excluding certain image domains, including those taken for accessibility needs such as descriptions for people with blindness~\citep{gurari2018vizwiz}.
Such biases in a benchmark, inherited from the multi- and unimodal datasets selected, serve the needs of English-speaking, able-bodied folks as a ``default.''

\section{Future Work}
\label{sec:future-work}
The initial findings from \climb\ reveal several promising opportunities and lines of research. 

\paragraph{Adapters} Primarily, we find that Adapters are effective at mitigating catastrophic forgetting, while achieving comparable performance to full model fine-tuning. 
However, our current Adapter experiments introduce an independent set of parameters for each multimodal task, which does not facilitate sharing of task knowledge between tasks. 
Within unimodal multi-task and continual learning, Hypernetworks~\citep{mahabadi2021parameter} and compositional Adapter modules~\citep{zhang2022continual} have been shown to facilitate knowledge transfer by generating Adapter parameters from shared task information. 
We plan to investigate how these methods generalize to multimodal CL, where shared information across tasks in either one or both modalities can influence generation of Adapter module parameters for new tasks.

\paragraph{Distribution Shifts with Multiple Modalities} Second, the stark performance degradation of the CL model after training on VCR, and the subsequent poor downstream few-shot transfer, invites study of how domain shifts in both vision and language inputs can affect upstream learning and forgetting, and can be mitigated.

\paragraph{Sequence-to-Sequence Tasks} Finally, as we noted in our Limitations, currently \climb\ only supports classification tasks. 
However, recently several ``generalist'' models have been developed, such as UnifiedIO~\citep{lu2022unified} and FLAVA~\citep{singh2022flava}, that can solve a large variety of multimodal and unimodal tasks by formulating all tasks as a Sequence-to-Sequence problem. 
We plan to extend \climb\ to support such all-purpose Sequence-to-Sequence models.

\begin{ack}
This work was supported by the Laboratory for Analytic Sciences (LAS), National Security \& Special Research Initiatives, and in part by DARPA under contract HR001121C0168.

\end{ack}
\clearpage

\bibliographystyle{plainnat}
\bibliography{references}

\clearpage

\section*{Checklist}

%%% BEGIN INSTRUCTIONS %%%
\begin{comment}
The checklist follows the references.  Please
read the checklist guidelines carefully for information on how to answer these
questions.  For each question, change the default \answerTODO{} to \answerYes{},
\answerNo{}, or \answerNA{}.  You are strongly encouraged to include a {\bf
justification to your answer}, either by referencing the appropriate section of
your paper or providing a brief inline description.  For example:
\begin{itemize}
  \item Did you include the license to the code and datasets? \answerYes{See Section~\ref{sec:cl}.}
  \item Did you include the license to the code and datasets? \answerNo{The code and the data are proprietary.}
  \item Did you include the license to the code and datasets? \answerNA{}
\end{itemize}
Please do not modify the questions and only use the provided macros for your
answers.  Note that the Checklist section does not count towards the page
limit.  In your paper, please delete this instructions block and only keep the
Checklist section heading above along with the questions/answers below.
\end{comment}
%%% END INSTRUCTIONS %%%

\begin{enumerate}

\item For all authors...
\begin{enumerate}
  \item Do the main claims made in the abstract and introduction accurately reflect the paper's contributions and scope?
    \answerYes{}
  \item Did you describe the limitations of your work?
    
    \answerYes{See Section~\ref{sec:limitations}}
  \item Did you discuss any potential negative societal impacts of your work?
    
    \answerYes{See Section~\ref{sec:limitations}}
  \item Have you read the ethics review guidelines and ensured that your paper conforms to them?
    \answerYes{}
\end{enumerate}

\item If you are including theoretical results...
\begin{enumerate}
  \item Did you state the full set of assumptions of all theoretical results?
    \answerNA{}
	\item Did you include complete proofs of all theoretical results?
    \answerNA{}
\end{enumerate}

\item If you ran experiments (e.g. for benchmarks)...
\begin{enumerate}
  \item Did you include the code, data, and instructions needed to reproduce the main experimental results (either in the supplemental material or as a URL)?
    
    \answerYes{Our code and instructions can be found at \url{https://github.com/GLAMOR-USC/CLiMB} (mentioned on Page 1).}
  \item Did you specify all the training details (e.g., data splits, hyperparameters, how they were chosen)?
    
    \answerYes{See Supplementary material.}
	\item Did you report error bars (e.g., with respect to the random seed after running experiments multiple times)?
	
	\answerYes{Yes, for language-only tasks (See Figure~\ref{fig:lang-only}), where we run experiments with three different random seeds. For other tasks, we do a single run with fixed the random seed, due to training time cost.}
    \item Did you include the total amount of compute and the type of resources used (e.g., type of GPUs, internal cluster, or cloud provider)?
    
    \answerYes{See Supplementary material.}
\end{enumerate}

\item If you are using existing assets (e.g., code, data, models) or curating/releasing new assets...
\begin{enumerate}
  \item If your work uses existing assets, did you cite the creators?
    \answerYes{}
  \item Did you mention the license of the assets?
    
    \answerYes{Our benchmark license is mentioned in the project's GitHub repository (linked above). Licenses of datasets included (but not distributed) are mentioned in the Supplementary material.}
  \item Did you include any new assets either in the supplemental material or as a URL?  \answerNA{}
  \item Did you discuss whether and how consent was obtained from people whose data you're using/curating?
    \answerNA{}
  \item Did you discuss whether the data you are using/curating contains personally identifiable information or offensive content?
    \answerNA{}
\end{enumerate}

\item If you used crowdsourcing or conducted research with human subjects...
\begin{enumerate}
  \item Did you include the full text of instructions given to participants and screenshots, if applicable?
    \answerNA{}
  \item Did you describe any potential participant risks, with links to Institutional Review Board (IRB) approvals, if applicable?
    \answerNA{}
  \item Did you include the estimated hourly wage paid to participants and the total amount spent on participant compensation?
    \answerNA{}
\end{enumerate}

\end{enumerate}

%%%%%%%%%%%%%%%%%%%%%%%%%%%%%%%%%%%%%%%%%%%%%%%%%%%%%%%%%%%%

\clearpage

\appendix

\section{Multimodal Task Details}
Table~\ref{tab:multimodal-task-details} shows details about individual multimodal tasks, including hyperparameters used to train \vilt\ for each task, and details about how low-shot versions of each task are sampled.

For NLVR2 and SNLI-VE, where the output labels are a small number of semantically meaningful categories (True/False and Entailment/Contradiction/Neutral respectively), we sample $N$ shots per output label to construct our low-shot training data. 
The 4 output labels in VCR are not semantically meaningful (since the options are interchangeable); hence, instead of sampling an equal number of training samples per label, we sample a percentage of the full training data instead. For VQAv2, the output label space is very large, and answers are not uniformly distributed across the training data, so instead of sampling $N$ shots per output label (answer) we again sample a percentage of the full VQAv2 training data.

\begin{table}[ht]
    \centering
    \small
    \begin{tabular}{lcccc}
    \toprule
        Task & VQAv2 & NLVR2 & SNLI-VE & VCR (Q $\rightarrow$ A) \\
    \midrule
    & \multicolumn{4}{c}{Task Details} \\
    \midrule
        Task Type & Classification & Classification    &  Classification & Multi-Choice  \\
        Visual Input & 1 Image & 2 Images    & 1 Image  & 1 Image, Object boxes \\
        Text Input & Question & Statement & Hypothesis  & 1 question, 4 answers  \\
        \# Output Labels & 3129 &  2  & 3  & 4  \\
        Random Score, $S_R^i$ (\%)  & 0.0 & 50.0 & 33.33 & 25.0  \\
    \midrule
    \multicolumn{5}{c}{Training Details/Hyperparameters} \\
    \midrule
        Learning Rate & $10^{-4}$ &  $10^{-4}$   &  $5 \times 10^{-5}$  &  $10^{-4}$  \\
        Weight Decay & $10^{-2}$ &  $10^{-2}$   &  $10^{-2}$  & $10^{-2}$ \\
        Adam Epsilon & $10^{-8}$ & $10^{-8}$ & $10^{-8}$   & $10^{-8}$   \\
        Num. Epochs & 10 & 10   & 5  & 10  \\
        Batch Size & 64 & 32   & 64  & 16  \\
    \midrule
    \multicolumn{5}{c}{Low-Shot Task Transformation} \\
    \midrule
        Number of shots per class & - & 2048 & 2048  & -   \\
        \% of training data & 5\% & 4.74\% & 1.16\%  & 5\%  \\
    \bottomrule
    \end{tabular}
    \caption{Task-specific implementation details}
    \label{tab:multimodal-task-details}
\end{table}

\section{\vilt\ Model Modification Details}
\subsection{Applying \vilt\ to Multi-Choice Tasks}

\subsubsection{Applying \vilt\ to VCR}

The VCR task provides object boxes, with each box corresponding to a grounded entity in the question. Unlike other pre-trained vision-language encoders~\citep{su2019vl,chen2019uniter} that use visual features from regions-of-interest (ROIs) in the image, \vilt\ is designed to operate over image patches, thus making it challenging to use the object box inputs provided in the VCR task. 
We follow previous work~\citep{zellers2021merlot,hessel2022abduction} and draw colored boxes directly on the image corresponding to grounded references in the text. The grounded text references, \textit{e.g.} \texttt{[person1], [car1]}, are replaced with gender-neutral names for persons and object class names for all other objects.
We use consistent mappings between the box colors and object names; for example, the \texttt{[person1]} object is always referenced with a green box in the image, and the name Casey in the text. 

During training and inference, each possible answer $a_i$ is paired with the question $q$, to form a sequence ``[CLS] $q$ [SEP] $a_i$''. Each question-answer option is passed into the \vilt\ transformer, and the classifier produces a scalar score for each choice on top of the [CLS] representation. The choice with the maximum score is selected as the answer.

\subsubsection{Applying \vilt\ to HellaSwag, PIQA, and CommonsenseQA}
The inputs of language-only multiple-choice tasks consist of two parts: a sentence $s$ (a sentence prefix in HellaSwag; a question in PIQA and CommonsenseQA), and a set of choices $A=\{a_1, a_2, ..., a_n\}$.
We follow the original implementations~\citep{zellers2019hellaswag, bisk2020piqa} to model these tasks, which consider different choices independently.
For each choice $a_i$, we concatenate $s$ and $a_i$ with special tokens as the input: "[CLS] $s$ [SEP] $a_i$ [SEP]".
We build the classifier, which outputs a scalar score for each choice, atop the [CLS] representation of \vilt\ transformer.
During fine-tuning, we aggregate the scores of different choices and train the model with cross-entropy loss over the choices. 

\subsection{Applying \vilt\ to Unimodal Tasks}
\paragraph{Sub-sampling.} We conduct low-shot experiments to test the model's transferability to unimodal tasks. 
However, different sub-samples the training set may lead to different results.
To deal with this issue, for every language-only task, we use three different random seeds for sub-sampling, leading to three different training subsets, and then report the mean and standard deviation of the accuracy scores on the full validation set.
For vision-only tasks, however, we observed low variances in accuracy across three sub-samplings ($39.35 \pm 0.4\%$ on Places 365; 16-shot per class).
Thus, we fix the random seed and only use a single training subset for vision-only tasks due to the computational cost.

\paragraph{What's the language input for vision-only tasks?} For vision-only tasks, we found that simply using "This is an image." as the language input works empirically well on all tasks.
While the performance could potentially be further improved by using more informative and contextualized textual inputs, we leave it as future work as this is not the primary focus of this paper.%we focus on continual learning in this paper.

\paragraph{What's the visual input for language-only tasks?} 
\begin{wrapfigure}[12]{r}{0.3\textwidth}
  \begin{center}
    \vspace{-12pt}
    \includegraphics[width=0.25\textwidth]{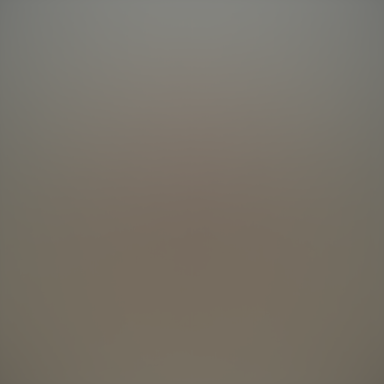}
  \end{center}
  \caption{The average image of the MS-COCO dataset.}
  \label{fig:coco_mean}
\end{wrapfigure}
For language-only tasks, to keep the visual input in-distribution, we first resize all MS-COCO training images into $384 \times 384$ and then average them into a single image (see Fig~\ref{fig:coco_mean}) as the vacuous visual input, since MS-COCO is one of the pre-training corpora of \vilt.
384 is the shorter-edge image size used by \vilt, and with the default $32 \times 32$ patch projection, it takes $(384/32) \times (384/32) = 144$ image tokens in the original implementation.
We also conduct ablation studies that include two baselines: (1) not inputting any image to \vilt\ at all, and (2) inputting the zero-vector image instead of the average image of the COCO dataset.
The first three rows in Table~\ref{tab:lang-sol} show that inputting the average image is slightly better than the other two baselines, presenting the benefit of using an in-distribution image, even when it is vacuous.

\paragraph{Language-and-vision token reallocation.} In language-only tasks, we would like to focus on the language inputs, which only accounts for 40 tokens in the original \vilt\ implementation, instead of the vacuous visual input, which accounts for 144 tokens.
Thus, we extend the language inputs by extending the positional embeddings to a maximum of 160, which is jointly learned during fine-tuning, and decrease the image tokens by downsample the image size to $128 \times 128$, which now only takes $(128/32) \times (128/32) = 16$ image tokens.
The last row in Table~\ref{tab:lang-sol} shows that this re-allocation of language-and-vision position embedding tokens notably improves the performance on language tasks.

\begin{table}[ht]
\centering
\begin{tabular}{l c c c}
\toprule
 & 16-shot & 32-shot & 128-shot\\
\midrule
ViLT-40 -no image & 51.2 ± 0.4 & 54.0 ± 1.2 & 56.8 ± 1.2\\
ViLT-40 -zero & 53.8 ± 0.1 & 54.8 ± 0.3 & 56.7 ± 0.7\\
ViLT-40 -avg & 53.7 ± 0.8 &  55.1 ± 1.0 & 58.0 ± 0.6\\
\hline
ViLT-160 -avg & \textbf{55.9} ± 2.1 & \textbf{57.8} ± 1.5 & \textbf{62.3} ± 0.5\\
\bottomrule
\end{tabular}
\caption{Accuracy (\%) of vacuous visual input variants on IMDb, with $N=\{16, 32, 128\}$ shot per class. $-l$ means the maximum language sequence length is $l$, where ViLT-160 -avg is the proposed method that reallocates the language-and-vision tokens and has fewer visual tokens than other rows.}
\label{tab:lang-sol}
\end{table}

\subsection{\viltbert\ Implementation Details}

The \viltbert\ model is a modification of the \vilt\ model that uses stronger language priors. Since the \vilt\ Transformer was initialized using weights from the vision transformer ViT~\citep{dosovitskiy2020image}, and pre-trained only on image caption datasets, the language understanding of \vilt\ is limited to a specific language domain of image captions, thus making it unsuitable for language-only tasks. We perform additional experiments with a \viltbert\ model~\citep{viltbert} that replaces the language input embeddings of the \vilt\ Transformer with language token representations extracted from a frozen, pre-trained BERT model. \viltbert\ has more effective language understanding ability due receiving inputs from BERT, but the more general language representations could hurt its performance on vision-language tasks.

\subsubsection{\vilt\ vs \viltbert\ Multimodal Task Comparison}

Table~\ref{tab:viltbert-multimodal} shows a comparison of pre-trained \vilt\ and \viltbert\ when directly trained on each of the upstream vision-language tasks. 
\viltbert\ underperforms \vilt\ across all multimodal tasks.
\begin{table}[h]
    \centering
    \begin{tabular}{lcccc}
        \toprule
         Model & VQAv2 & NLVR2 & SNLI-VE & VCR \\
        \midrule
        \vilt & 67.70\% & 73.07\% & 76.31\% & 61.31\% \\
         \viltbert & 65.80\% & 65.57\% & 74.12\% & 59.46\% \\
        \bottomrule
    \end{tabular}
    \caption{Comparison of pre-trained \vilt\ versus \viltbert\ when trained directly on each of the upstream multimodal tasks. \viltbert\ consistently underperforms \vilt's accuracy.}
    \label{tab:viltbert-multimodal}
\end{table}

\subsubsection{\vilt\ vs \viltbert\ Language-Only Task Comparison}
Table~\ref{table:lang-viltbert} compares pre-trained \vilt\ and \viltbert\ when directly fine-tuned on downstream language-only tasks, showing that \viltbert\ can significantly improve the accuracy over \vilt.

\begin{table}[h]
  \centering
  \begin{tabular}{lcccc}
    \toprule
     \multirow{2}{*}{Model} & \multicolumn{2}{c}{IMDb} & \multicolumn{2}{c}{SST-2}  \\
     \cmidrule(lr){2-3}\cmidrule(lr){4-5}
     & 16 & 32 & 16 & 32 \\
     \midrule
     \vilt & 55.9 ± 2.1 & 57.8 ± 1.5 & 57.8 ± 3.6 & 58.5 ± 7.4\\
     \viltbert & \textbf{64.8} ± 2.0 & \textbf{70.0} ± 1.7 & \textbf{59.9} ± 3.0 & \textbf{67.0} ± 3.2 \\
    \bottomrule
    \end{tabular}
    
    \phantom{0}
    
    \begin{tabular}{lcccccc}
    \toprule
     \multirow{2}{*}{Model} & \multicolumn{2}{c}{HellaSwag} & \multicolumn{2}{c}{CommonsenseQA} & \multicolumn{2}{c}{PIQA}\\
     \cmidrule(lr){2-3}\cmidrule(lr){4-5}\cmidrule(lr){6-7}
     & 1024 & 4096 & 1024 & 4096 & 1024 & 4096\\
     \midrule
     \vilt & 26.5 ± 0.3 & 27.7 ± 0.3 & 23.0 ± 2.0 & 26.3 ± 0.5 & 52.0 ± 0.7 & 54.8 ± 0.5\\
     \viltbert & \textbf{31.7} ± 0.5 & \textbf{32.9} ± 0.8 & \textbf{41.8} ± 0.6 & \textbf{43.4} ± 0.6 & \textbf{54.6} ± 1.0 & \textbf{58.0} ± 0.8\\
    \bottomrule
    \end{tabular}
    
    \caption{Comparisons between \vilt\ and \viltbert\ on downstream language-only tasks. Each value is the average accuracy ($\%$) and standard deviation over 3 runs. We experiment with $\{16, 32\}$-shot per class on IMDB, SST-2 and sub-sample $\{1024, 4096\}$ training data for HellaSwag, CommonSenseQA, and PIQA. \viltbert\ consistently achieves higher accuracy than \vilt.}
  \label{table:lang-viltbert}
\end{table}

\begin{figure}[t]
    \centering
    \includegraphics[width=1\linewidth]{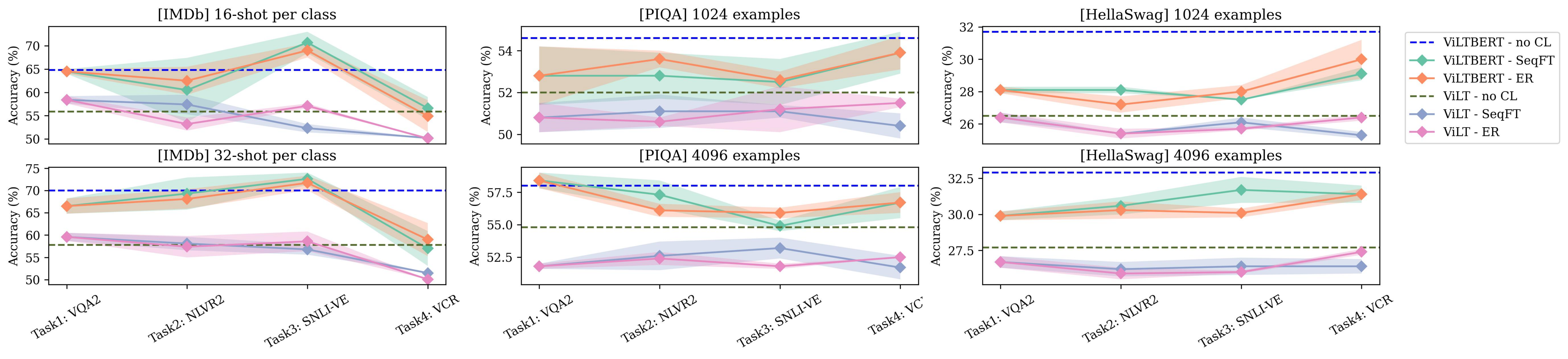}
    \caption{
        Comparisons between \vilt\ and \viltbert\ with checkpoints from different CL algorithms on downstream language-only tasks.
        We conduct three runs of different training sub-samplings and plot the absolute accuracy with shaded standard deviation.
    }
    \label{fig:lang-only-supp}
\end{figure}

Similarly, Figure~\ref{fig:lang-only-supp} shows that \viltbert\ strictly improves absolute accuracy over \vilt\ in direct fine-tuning and CL settings for downstream language-only tasks.

\section{Algorithm Implementation Details}

For \textbf{Sequential Fine-tuning}, we fine-tune the shared encoder parameters when learning each task, whereas for \textbf{Frozen Encoder}, we keep the shared encoder frozen and only fine-tune the task-specific classification layers. In \textbf{Frozen Bottom-K}, the embedding parameters and the bottom $K$ (< 12) transformer layers are frozen; we set $K$=9 in our experiments.

The \textbf{Experience Replay (ER)} algorithm has two hyperparameters: the percentage of each task's training samples to be stored in the memory buffer, and the frequency with which to perform a ``replay step''. We set these hyperparameters as 1\% of training data and 100 training steps, respectively. We sample training examples for the memory buffer randomly from the training dataset; alternatives include sampling an equal number of training examples per output class.

\textbf{Elastic Weight Consolidation (EWC)} consists of computing the Fisher information matrix from the training data, which determines the importance of each parameter in the shared encoder. Rather than doing a full pass through the whole training data to construct the Fisher information matrix for each task, we use only 1\% of the training examples. During training an upstream task, we sample one of the previous tasks and compute the L2 loss between parameter values in the current encoder and the previous task's encoder checkpoint. The L2 loss is weighted by that parameter's Fisher information and summed across all parameters. This EWC loss $\mathcal{L}_{EWC}$ is multiplied by a constant
$\lambda$ and added to the upstream task loss $\mathcal{L}_{task}$. We select $\lambda = 10^2$ based on a hyperparameter sweep.

\textbf{Adapters} add a 2-layer MLP, also known as an Adapter module, after every Self-Attention and Feed-forward layer in each Transformer block. Following the original Houlsby configuration~\citep{houlsby2019parameter}, the first layer of each Adapter module is a downsampling layer, which reduces the dimensionality of the input features by a factor of 16, followed by a GELU activation function, and finally an upsampling layer which produces an output representation with the same dimensionality as the Adapter input.

\section{Full \vilt\ Results}
\subsection{Full Catastrophic Forgetting Results}
In Table~\ref{tab:full-forgetting}, we present full forgetting transfer numbers for all six CL algorithms, which were reported in a compact form in Figure~\ref{subfig:forgetting}.

\begin{table}[ht]
    \setlength{\aboverulesep}{0pt}
    \setlength{\belowrulesep}{0pt}
    \centering
    \begin{small}
\begin{tabular}{lrrr}
    \multicolumn{4}{c}{CL Algorithm: Sequential Fine-tuning} \\
    \toprule
    %\multicolumn{7}{c}{Comparison of Algorithms (Fixed Task Order, Encoder=ViLT)}\\
    %\toprule
    \multirow{2}{*}{\backslashbox{Checkpoint}{Evaluated on}} & \multicolumn{1}{c}{Task 1} & \multicolumn{1}{c}{Task 2} & \multicolumn{1}{c}{Task 3} \\
    &  \multicolumn{1}{c}{VQAv2} & \multicolumn{1}{c}{NLVR2} & \multicolumn{1}{c}{SNLI-VE} \\
    \midrule
    After training on that task & [67.79]	& [72.66] & [74.89] \\
    \midrule
    Task 2: NLVR2  &  \transfer{40.97}{40.02}	& - & - \\
    Task 3: SNLI-VE &   \transfer{39.25}{41.18} & \transfer{43.81}{62.73} & - \\
    Task 4: VCR &  \transfer{63.90}{24.47} & \transfer{93.74}{51.24} & \transfer{89.93}{37.52} \\
    \bottomrule
    \\
    \multicolumn{4}{c}{CL Algorithm: Frozen Encoder} \\
    \toprule
    %\multicolumn{7}{c}{Comparison of Algorithms (Fixed Task Order, Encoder=ViLT)}\\
    %\toprule
    \multirow{2}{*}{\backslashbox{Checkpoint}{Evaluated on}} & \multicolumn{1}{c}{Task 1} & \multicolumn{1}{c}{Task 2} & \multicolumn{1}{c}{Task 3} \\
    &  \multicolumn{1}{c}{VQAv2} & \multicolumn{1}{c}{NLVR2} & \multicolumn{1}{c}{SNLI-VE} \\
    \midrule
    After training on that task & [58.15]	& [63.66] & [69.45] \\
    \midrule
    Task 2: NLVR2  &  \transfer{-0.38}{58.37}	& - & - \\
    Task 3: SNLI-VE &   \transfer{-0.38}{58.37} & \transfer{-0.31}{63.70} & - \\
    Task 4: VCR &  \transfer{-0.38}{58.37} & \transfer{-0.42}{63.72} & \transfer{0.00}{69.45} \\
    \bottomrule
    \\
    \multicolumn{4}{c}{CL Algorithm: Frozen Bottom-9} \\
    \toprule
    %\multicolumn{7}{c}{Comparison of Algorithms (Fixed Task Order, Encoder=ViLT)}\\
    %\toprule
    \multirow{2}{*}{\backslashbox{Checkpoint}{Evaluated on}} & \multicolumn{1}{c}{Task 1} & \multicolumn{1}{c}{Task 2} & \multicolumn{1}{c}{Task 3} \\
    &  \multicolumn{1}{c}{VQAv2} & \multicolumn{1}{c}{NLVR2} & \multicolumn{1}{c}{SNLI-VE} \\
    \midrule
    After training on that task & [67.30]	& [72.94] & [74.90] \\
    \midrule
    Task 2: NLVR2  &  \transfer{16.97}{55.90}	& - & - \\
    Task 3: SNLI-VE &   \transfer{21.36}{52.93} & \transfer{29.32}{66.21} & - \\
    Task 4: VCR &  \transfer{71.61}{19.11} & \transfer{78.52}{54.93} & \transfer{35.01}{60.34} \\
    \bottomrule
    \\
    \multicolumn{4}{c}{CL Algorithm: Experience Replay} \\
    \toprule
    %\multicolumn{7}{c}{Comparison of Algorithms (Fixed Task Order, Encoder=ViLT)}\\
    %\toprule
    \multirow{2}{*}{\backslashbox{Checkpoint}{Evaluated on}} & \multicolumn{1}{c}{Task 1} & \multicolumn{1}{c}{Task 2} & \multicolumn{1}{c}{Task 3} \\
    &  \multicolumn{1}{c}{VQAv2} & \multicolumn{1}{c}{NLVR2} & \multicolumn{1}{c}{SNLI-VE} \\
    \midrule
    After training on that task & [67.87]	& [73.20] & [75.08] \\
    \midrule
    Task 2: NLVR2  &  \transfer{12.88}{59.13}	& - & - \\
    Task 3: SNLI-VE &   \transfer{12.96}{59.07} & \transfer{17.10}{69.23} & - \\
    Task 4: VCR &  \transfer{43.62}{38.27} & \transfer{78.27}{55.04} & \transfer{33.45}{61.11} \\
    \bottomrule
    \\
    \multicolumn{4}{c}{CL Algorithm: Elastic Weight Consolidation} \\
    \toprule
    %\multicolumn{7}{c}{Comparison of Algorithms (Fixed Task Order, Encoder=ViLT)}\\
    %\toprule
    \multirow{2}{*}{\backslashbox{Checkpoint}{Evaluated on}} & \multicolumn{1}{c}{Task 1} & \multicolumn{1}{c}{Task 2} & \multicolumn{1}{c}{Task 3} \\
    &  \multicolumn{1}{c}{VQAv2} & \multicolumn{1}{c}{NLVR2} & \multicolumn{1}{c}{SNLI-VE} \\
    \midrule
    After training on that task & [67.84] &	[72.39]	& [74.38] \\
    \midrule
    Task 2: NLVR2  &  \transfer{39.81}{40.83}	& - & - \\
    Task 3: SNLI-VE &   \transfer{31.52}{46.46} & \transfer{25.73}{66.66} & - \\
    Task 4: VCR &  \transfer{65.25}{23.58} & \transfer{81.03}{54.25} & \transfer{73.61}{43.34} \\
    \bottomrule
    \\
    \multicolumn{4}{c}{CL Algorithm: Adapters} \\
    \toprule
    %\multicolumn{7}{c}{Comparison of Algorithms (Fixed Task Order, Encoder=ViLT)}\\
    %\toprule
    \multirow{2}{*}{\backslashbox{Checkpoint}{Evaluated on}} & \multicolumn{1}{c}{Task 1} & \multicolumn{1}{c}{Task 2} & \multicolumn{1}{c}{Task 3} \\
    &  \multicolumn{1}{c}{VQAv2} & \multicolumn{1}{c}{NLVR2} & \multicolumn{1}{c}{SNLI-VE} \\
    \midrule
    After training on that task & [68.10]	& [73.66] & [76.08] \\
    \midrule
    Task 2: NLVR2  &  \transfer{-0.01}{68.11}	& - & - \\
    Task 3: SNLI-VE &   \transfer{0.04}{68.07} & \transfer{3.51}{72.83} & - \\
    Task 4: VCR &  \transfer{0.67}{67.64} & \transfer{6.48}{72.13} & \transfer{0.89}{75.70} \\
    \bottomrule
    \end{tabular}
    \end{small}
    \caption{Full numbers for forgetting transfer $\forgetting{j}{i}$ of previously seen tasks for each CL algorithm. We also show the  transfer score $[S_\algorithm^{j \leftarrow i}]$ when evaluated on that task after training on future task $i$. The first row contains task score $[S_\algorithm^j]$ after originally training on $j^{th}$ task.
    }
\label{tab:full-forgetting}
\end{table}

\subsection{Full Results of Different Upstream Task Orders}
\begin{table}[ht]

\begin{subtable}[h]{\textwidth}
        \setlength{\aboverulesep}{0pt}
    \setlength{\belowrulesep}{0pt}
    \centering
    \begin{small}
    \begin{tabular}{rrrr}
    \multicolumn{4}{c}{Directly fine-tuning pre-trained \vilt\ on each task}\\
    \toprule
    \multicolumn{1}{c}{VQAv2} & \multicolumn{1}{c}{NLVR2} & \multicolumn{1}{c}{SNLI-VE} & \multicolumn{1}{c}{VCR} \\
    \midrule
    \multicolumn{1}{c}{[67.70]} & \multicolumn{1}{c}{[73.07]} & \multicolumn{1}{c}{[76.31]} & \multicolumn{1}{c}{[61.31]} \\
    \bottomrule
    \\
    \multicolumn{4}{c}{Task Order: VQAv2 $\rightarrow$ NLVR2 $\rightarrow$ SNLI-VE $\rightarrow$ VCR}\\
    \toprule
    \multicolumn{1}{c}{Task 1} & \multicolumn{1}{c}{Task 2} & \multicolumn{1}{c}{Task 3} & \multicolumn{1}{c}{Task 4} \\
    \multicolumn{1}{c}{VQAv2} & \multicolumn{1}{c}{NLVR2} & \multicolumn{1}{c}{SNLI-VE} & \multicolumn{1}{c}{VCR} \\
    \midrule
    \transfer{0.13}{67.79}	& \transfer{-1.80}{72.66} & \transfer{-3.33}{74.89} & \transfer{-5.09}{59.47} \\
    \bottomrule
    \\
    \multicolumn{4}{c}{Task Order: SNLI-VE $\rightarrow$ VCR $\rightarrow$ VQAv2 $\rightarrow$ NLVR2}\\
    \toprule
    \multicolumn{1}{c}{Task 1} & \multicolumn{1}{c}{Task 2} & \multicolumn{1}{c}{Task 3} & \multicolumn{1}{c}{Task 4} \\
    \multicolumn{1}{c}{SNLI-VE} & \multicolumn{1}{c}{VCR} & \multicolumn{1}{c}{VQAv2} & \multicolumn{1}{c}{NLVR2} \\
    \midrule
    \transfer{-0.07}{76.29}	& \transfer{-1.55}{60.75} & \transfer{-6.55}{63.27} & \transfer{-21.35}{67.65} \\
    \bottomrule
    \\
    \multicolumn{4}{c}{Task Order: NLVR2 $\rightarrow$ VQAv2 $\rightarrow$ VCR $\rightarrow$ SNLI-VE}\\
    \toprule
    \multicolumn{1}{c}{Task 1} & \multicolumn{1}{c}{Task 2} & \multicolumn{1}{c}{Task 3} & \multicolumn{1}{c}{Task 4} \\
    \multicolumn{1}{c}{NLVR2} & \multicolumn{1}{c}{VQAv2} & \multicolumn{1}{c}{VCR} & \multicolumn{1}{c}{SNLI-VE} \\
    \midrule
    \transfer{0.06}{73.25}	& \transfer{-1.52}{66.55} & \transfer{-6.03}{59.10} & \transfer{-7.88}{73.07} \\
    \bottomrule
    
    \end{tabular}
    \end{small}
    \caption{Full knowledge transfer results with different task orders.
    }
    \label{subtab:taskorder-ukt}
\end{subtable}

\begin{subtable}[h]{\textwidth}
    \setlength{\aboverulesep}{0pt}
    \setlength{\belowrulesep}{0pt}
    \centering
    \begin{small}
\begin{tabular}{lrrr}
    \multicolumn{4}{c}{Task Order: VQAv2 $\rightarrow$ NLVR2 $\rightarrow$ SNLI-VE $\rightarrow$ VCR} \\
    \toprule
    %\multicolumn{7}{c}{Comparison of Algorithms (Fixed Task Order, Encoder=ViLT)}\\
    %\toprule
    \multirow{2}{*}{\backslashbox{Checkpoint}{Evaluated on}} & \multicolumn{1}{c}{Task 1} & \multicolumn{1}{c}{Task 2} & \multicolumn{1}{c}{Task 3} \\
    &  \multicolumn{1}{c}{VQAv2} & \multicolumn{1}{c}{NLVR2} & \multicolumn{1}{c}{SNLI-VE} \\
    \midrule
    After training on that task & [67.79]	& [72.66] & [74.89] \\
    \midrule
    Task 2: NLVR2  &  \transfer{40.97}{40.02}	& - & - \\
    Task 3: SNLI-VE &   \transfer{39.25}{41.18} & \transfer{43.81}{62.73} & - \\
    Task 4: VCR &  \transfer{63.90}{24.47} & \transfer{93.74}{51.24} & \transfer{89.93}{37.52} \\
    \bottomrule
    \\
    \multicolumn{4}{c}{Task Order: SNLI-VE $\rightarrow$ VCR $\rightarrow$ VQAv2 $\rightarrow$ NLVR2} \\
    \toprule
    %\multicolumn{7}{c}{Comparison of Algorithms (Fixed Task Order, Encoder=ViLT)}\\
    %\toprule
    \multirow{2}{*}{\backslashbox{Checkpoint}{Evaluated on}} & \multicolumn{1}{c}{Task 1} & \multicolumn{1}{c}{Task 2} & \multicolumn{1}{c}{Task 3} \\
    &  \multicolumn{1}{c}{SNLI-VE} & \multicolumn{1}{c}{VCR} & \multicolumn{1}{c}{VQAv2} \\
    \midrule
    After training on that task & [76.29]	& [60.75] & [63.27] \\
    \midrule
    Task 2: VCR  &  \transfer{84.50}{39.99}	& - & - \\
    Task 3: VQAv2 &   \transfer{85.86}{39.40} & \transfer{91.47}{28.05} & - \\
    Task 4: NLVR2 &  \transfer{77.56}{42.97} & \transfer{86.11}{29.97} & \transfer{41.94}{36.73} \\
    \bottomrule
    \\
    \multicolumn{4}{c}{Task Order: NLVR2 $\rightarrow$ VQAv2 $\rightarrow$ VCR $\rightarrow$ SNLI-VE} \\
    \toprule
    %\multicolumn{7}{c}{Comparison of Algorithms (Fixed Task Order, Encoder=ViLT)}\\
    %\toprule
    \multirow{2}{*}{\backslashbox{Checkpoint}{Evaluated on}} & \multicolumn{1}{c}{Task 1} & \multicolumn{1}{c}{Task 2} & \multicolumn{1}{c}{Task 3} \\
    &  \multicolumn{1}{c}{NLVR2} & \multicolumn{1}{c}{VQAv2} & \multicolumn{1}{c}{VCR} \\
    \midrule
    After training on that task & [73.25]	& [66.55] & [59.10] \\
    \midrule
    Task 2: VQAv2  &  \transfer{58.06}{59.68}	& - & - \\
    Task 3: VCR &   \transfer{90.63}{52.16} & \transfer{68.69}{20.87} & - \\
    Task 4: SNLI-VE &  \transfer{91.75}{51.90} & \transfer{62.59}{24.94} & \transfer{34.04}{47.51} \\
    \bottomrule
    \end{tabular}
    \end{small}
    \caption{Full forgetting results with different task orders.
    }
    \label{subtab:taskorder-forgetting}
\end{subtable}
\caption{Effects of CL task order on the \vilt\ encoder's upstream knowledge transfer and forgetting. }
\label{tab:full-taskorder}
\end{table}

Table~\ref{subtab:taskorder-ukt} contains full results of upstream knowledge transfer $\knowledgetransfer{i}$,when the \vilt\ encoder sees different sequences of upstream tasks. We use Sequential Fine-tuning for all these experiments. Table~\ref{subtab:taskorder-forgetting} shows the forgetting of previous tasks, for these different upstream task orders. We previously summarized these results visually in Figure~\ref{subfig:taskorder}.

\subsection{Full Results of Low-Shot Multimodal Transfer}
\begin{table}[ht]
    \setlength{\aboverulesep}{0pt}
    \setlength{\belowrulesep}{0pt}
    \centering
    \begin{small}
\begin{tabular}{lrrr}
    \multicolumn{4}{c}{Directly fine-tuning on each task} \\
    \toprule
    %\multicolumn{7}{c}{Comparison of Algorithms (Fixed Task Order, Encoder=ViLT)}\\
    %\toprule
     & \multicolumn{1}{c}{Task 2} & \multicolumn{1}{c}{Task 3} & \multicolumn{1}{c}{Task 4} \\
    Training Data presented &  \multicolumn{1}{c}{NLVR2} & \multicolumn{1}{c}{SNLI-VE} & \multicolumn{1}{c}{VCR} \\
    \midrule
    Full Training Data, $S_{PT}^i$  &  [73.07] & [76.31] & [61.31] \\
    Low-Shot Transfer, $S_{PT}^{LS(i)}$ &   [62.46] & [65.67] & [43.23] \\
    \\
    \multicolumn{4}{c}{CL Algorithm: Sequential Fine-tuning} \\
    \toprule
    %\multicolumn{7}{c}{Comparison of Algorithms (Fixed Task Order, Encoder=ViLT)}\\
    %\toprule
    \multicolumn{1}{r}{Low-Shot Transfer to} & \multicolumn{1}{c}{Task 2} & \multicolumn{1}{c}{Task 3} & \multicolumn{1}{c}{Task 4} \\
    &  \multicolumn{1}{c}{NLVR2} & \multicolumn{1}{c}{SNLI-VE} & \multicolumn{1}{c}{VCR} \\
    \midrule
    After training on Task 1: VQAv2  &  \transfer{-8.19}{61.44} & \transfer{-4.51}{64.21} & \transfer{-13.71}{40.73} \\
    After training on Task 2: NLVR2 &   - & \transfer{-14.87}{60.86} & \transfer{-26.60}{38.48} \\
    After training on Task 3: SNLI-VE &  - & - & \transfer{-18.71}{39.82} \\
    \bottomrule
    \\
    \multicolumn{4}{c}{CL Algorithm: Frozen Bottom-9} \\
    \toprule
    %\multicolumn{7}{c}{Comparison of Algorithms (Fixed Task Order, Encoder=ViLT)}\\
    %\toprule
    \multicolumn{1}{r}{Low-Shot Transfer to} & \multicolumn{1}{c}{Task 2} & \multicolumn{1}{c}{Task 3} & \multicolumn{1}{c}{Task 4} \\
    &  \multicolumn{1}{c}{NLVR2} & \multicolumn{1}{c}{SNLI-VE} & \multicolumn{1}{c}{VCR} \\
    \midrule
    After training on Task 1: VQAv2  &  \transfer{-15.73}{60.50} & \transfer{-0.87}{65.39} & \transfer{-4.22}{42.46} \\
    After training on Task 2: NLVR2 &   - & \transfer{-0.99}{65.35} & \transfer{-10.48}{41.32} \\
    After training on Task 3: SNLI-VE &  - & - & \transfer{-4.00}{42.50} \\
    \bottomrule
    \\
    \multicolumn{4}{c}{CL Algorithm: Experience Replay} \\
    \toprule
    %\multicolumn{7}{c}{Comparison of Algorithms (Fixed Task Order, Encoder=ViLT)}\\
    %\toprule
    \multicolumn{1}{r}{Low-Shot Transfer to} & \multicolumn{1}{c}{Task 2} & \multicolumn{1}{c}{Task 3} & \multicolumn{1}{c}{Task 4} \\
    &  \multicolumn{1}{c}{NLVR2} & \multicolumn{1}{c}{SNLI-VE} & \multicolumn{1}{c}{VCR} \\
    \midrule
    After training on Task 1: VQAv2  &  \transfer{-7.95}{61.47} & \transfer{-1.76}{65.10} & \transfer{-15.47}{40.41} \\
    After training on Task 2: NLVR2 &   - & \transfer{-10.48}{62.28} & \transfer{-26.82}{38.34} \\
    After training on Task 3: SNLI-VE &  - & - & \transfer{-18.38}{39.88} \\
    \bottomrule
    \\
    \multicolumn{4}{c}{CL Algorithm: Elastic Weight Consolidation} \\
    \toprule
    %\multicolumn{7}{c}{Comparison of Algorithms (Fixed Task Order, Encoder=ViLT)}\\
    %\toprule
    \multicolumn{1}{r}{Low-Shot Transfer to} & \multicolumn{1}{c}{Task 2} & \multicolumn{1}{c}{Task 3} & \multicolumn{1}{c}{Task 4} \\
    &  \multicolumn{1}{c}{NLVR2} & \multicolumn{1}{c}{SNLI-VE} & \multicolumn{1}{c}{VCR} \\
    \midrule
    After training on Task 1: VQAv2  &  \transfer{-13.24}{60.81} & \transfer{-2.01}{65.02} & \transfer{-15.52}{40.40} \\
    After training on Task 2: NLVR2 &   - & \transfer{-17.53}{60.00} & \transfer{-29.29}{37.89} \\
    After training on Task 3: SNLI-VE &  - & - & \transfer{-22.87}{39.06} \\
    \bottomrule
    \end{tabular}
    \end{small}
    \caption{Full low-shot multiodal transfer results, when transferring \vilt\ checkpoints from upstream CL training to future multimodal tasks.
    }
\label{tab:full-lowshotmultimodal}
\end{table}

In Table~\ref{tab:full-lowshotmultimodal}, we present the full results when the CL-learned \vilt\ encoder, after training on the $i^{th}$ task, is trained on future low-shot tasks $\task_{VL}^{LS(j)}$ for $j = \{i+1,...,K_{VL}\}$. The first section of the table contains a comparison of \vilt's performance when directly fine-tuned on each task, when both full training data and low-shot versions of the task are available. The following sections show the low-shot transfer when upstream checkpoints, trained using four of our six CL algorithms, are used to fine-tuned on low-shot tasks. We do not perform experiments with the Frozen-Encoder and Adapter algorithms, as the encoder parameters are identical to the pre-trained checkpoint.

\subsection{Full Results of Low-Shot Unimodal Transfer}
\begin{figure}[ht]
    \centering
    \includegraphics[width=1\linewidth]{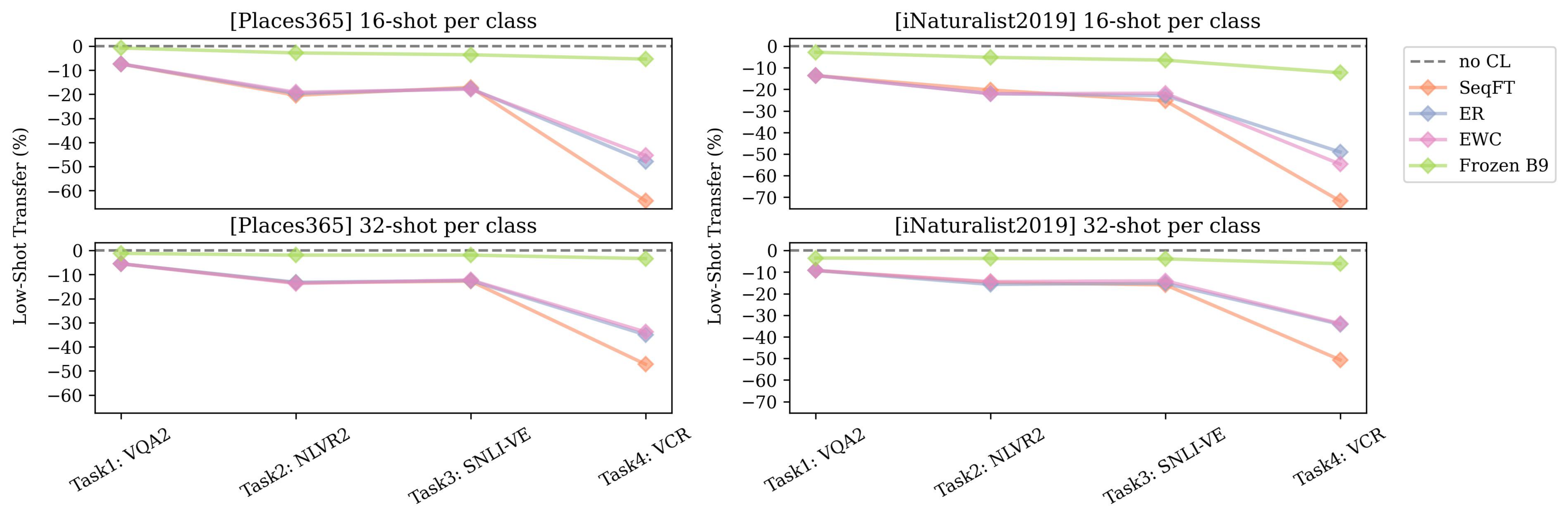}
    \caption{
        Low-Shot Transfer (\%) comparison between different CL algorithms on downstream vision-only tasks (left: Places365; right: iNaturalist2019).
    }
    \label{fig:vision-only-supp}
\end{figure}

\paragraph{Vision-only downstream tasks.} Table~\ref{table:full-v} presents the full results of vision-only tasks in absolute accuracy (\%).
Figure~\ref{fig:vision-only-supp} plots the same results with Low-Shot Transfer (\%).
First, in single-task fine-tuning, we only include a single task in the upstream phase and compare the influence of different upstream tasks to downstream low-shot transfer.
We found that across all vision-only downstream tasks, SNLI-VE > VQAv2 > NLVR2 > VCR, where VCR as the upstream task significantly damages the model performance.
Second, current CL algorithms always \emph{hurt} low-shot transfer compared to direct fine-tuning.
Among them, Frozen Bottom-9 is the least harmful algorithm.
Experience Replay and EWC perform similarly, and both are notably better than Sequential Fine-Tuning after training on VCR.
In conclusion, the pre-trained \vilt\ already achieves decent performance on low-shot vision-only classification tasks.
Meanwhile, with current CL algorithms, the model does not benefit from training on more vision-and-language upstream tasks, but suffers from forgetting useful visual representations, learned in pretraining, for downstream tasks.

\begin{table}[ht]
  \centering
  \begin{tabular}{lcccccccc}
    \toprule
     \multirow{2}{*}{\backslashbox{Checkpoint}{Task}} & \multicolumn{2}{c}{ImageNet} & \multicolumn{2}{c}{iNat2019} & \multicolumn{2}{c}{Places365} &
     \multicolumn{2}{c}{COCO} \\
     \cmidrule(lr){2-3}\cmidrule(lr){4-5}\cmidrule(lr){6-7}\cmidrule(lr){8-9}
     & 16 & 32 & 16 & 32 & 16 & 32 & 5\% & 10\% \\
     \midrule
     \emph{Direct Fine-Tuning} \\
     \cmidrule(lr){1-1}
     \vilt & 64.4 & 67.7 & 46.3 & 54.1 & 39.2 & 41.7 & 77.1 & 78.5\\
     \midrule
     \emph{CL: Singe-Task Fine-Tuning} \\
     \cmidrule(lr){1-1}
     After Task1: SNLI-VE & 62.3 & 66.3 & 43.6 & 53.1 & 37.6 & 40.5 & 74.6 & 77.4\\
     After Task1: VQAv2 & 58.8 & 63.3 & 40.0 & 49.1 & 36.3 & 39.4 & 73.2 & 75.7\\
     After Task1: NLVR2 & 56.2 & 62.7 & 36.4 & 48.4 & 31.9 & 37.0 & 67.3 & 73.1\\
     After Task1: VCR & 25.2 & 45.8 & 10.3 & 34.4 & 17.6 & 26.6 & 60.7 & 66.8\\
     \midrule
     \emph{CL: Sequential Fine-Tuning} \\
     \cmidrule(lr){1-1}
     After Task2: NLVR2 & 59.0 & 50.8 & 36.9 & 46.1 & 31.2 & 36.1 & 68.7 & 72.3\\
     After Task3: SNLI-VE & 51.5 & 59.1 & 34.6 & 45.5 & 32.5 & 36.4 & 70.3 & 72.6\\
     After Task4: VCR & 17.3 & 33.1 & 13.1 & 26.7 & 14.0 & 22.0 & 55.1 & 62.0\\
     \midrule
     \emph{CL: Experience Replay} \\
     \cmidrule(lr){1-1}
     After Task2: NLVR2 & 52.0 & 59.1 & 36.1 & 45.6 & 31.5 & 36.2 & 70.1 & 72.4\\
     After Task3: SNLI-VE & 52.2 & 58.8 & 35.7 & 45.9 & 32.3 & 36.5 & 70.6 & 72.9\\
     After Task4: VCR & 31.6 & 45.0 & 23.6 & 35.6 & 20.4 & 27.1 & 59.7 & 65.9\\  
     \midrule
     \emph{CL: EWC} \\
     \cmidrule(lr){1-1}
     After Task2: NLVR2 & 52.6 & 59.6 & 36.1 & 46.3 & 31.7 & 36.0 & 69.5 & 72.6\\
     After Task3: SNLI-VE & 52.9 & 59.2 & 36.2 & 46.5 & 32.2 & 36.6 & 70.2 & 73.0\\
     After Task4: VCR & 30.4 & 45.0 & 21.0 & 35.8 & 21.4 & 27.6 & 60.4 & 65.6\\   
     \midrule
     \emph{CL: Frozen Bottom-9} \\
     \cmidrule(lr){1-1}
     After Task1: VQAv2 & 62.8 & 66.3 & 45.0 & 52.2 & 38.9 & 41.2 & 76.6 & 78.1\\
     After Task2: NLVR2 & 62.2 & 66.0 & 43.9 & 52.1 & 38.1 & 40.9 & 76.1 & 78.1\\
     After Task3: SNLI-VE & 62.0 & 65.8 & 43.3 & 52.0 & 37.8 & 40.9 & 75.9 & 77.9\\
     After Task4: VCR & 60.2 & 65.1 & 40.6 & 50.8 & 37.1 & 40.3 & 75.4 & 77.8\\
     \bottomrule
    \end{tabular}
    
    \caption{Comparisons between different CL algorithms on vision-only tasks. We experiment with $\{16, 32\}$-shot per class on ImageNet-1000, iNaturalist 2019, and Places 365 datasets. For COCO multi-label object detection task, we sub-sample $\{5\%, 10\%\}$ of the training data. All CL algorithms hurt the accuracy (\%) compared to direct fine-tuning, while Frozen Bottom-9 is the least harmful one. Comparing different upstream tasks, SNLI-VE > VQAv2 > NLVR2 > VCR across all four downstream tasks, where VCR greatly damages the performance. Note that for Sequential Fine-Tuning, Experience Replay, and EWC, the result of "After Task1: VQAv2" is shown under  Single-Task Fine-Tuning, as there are no differences between these CL algorithms in the first task.}
  \label{table:full-v}
\end{table}

\paragraph{Language-only downstream tasks.}
Table~\ref{table:full-l} presents the full results of language-only tasks in absolute accuracy (\%).
First, \vilt\ performs poorly on language-only tasks.
Similar to our findings in vision-only tasks and multimodal tasks, including VCR as one of the upstream tasks hurts the model performance on downstream tasks, most notably on SST-2 and IMDb.
Although VCR is also a multiple-choice commonsense reasoning task, it does not benefit HellaSwag, CommonsenseQA, and PIQA.
On the other hand, continual learning sometimes improves the accuracy, especially on SST-2 and IMDb.
\climb\ facilitates further investigation into these phenomena.
\begin{table}[ht]
  \small
  \centering
  \begin{tabular}{lcccccccccc}
    \toprule
     \multirow{2}{*}{\backslashbox{Checkpoint}{Task}} & \multicolumn{2}{c}{IMDb} & \multicolumn{2}{c}{SST-2} & \multicolumn{2}{c}{HellaSwag} &
     \multicolumn{2}{c}{ComQA} &
     \multicolumn{2}{c}{PIQA}\\
     \cmidrule(lr){2-3}\cmidrule(lr){4-5}\cmidrule(lr){6-7}\cmidrule(lr){8-9}\cmidrule(lr){10-11}
     & 16 & 32 & 16 & 32 & 1024 & 4096 & 1024 & 4096 & 1024 & 4096 \\
     \midrule
     \emph{Direct Fine-Tuning} \\
     \cmidrule(lr){1-1}
     \vilt & 55.9 & 57.8 & 57.8 & 58.5 & 26.5 & 27.7 & 23.0 & 26.3 & 52.0 & 54.8\\
     \midrule
     \emph{CL: Singe-Task Fine-Tuning} \\
     \cmidrule(lr){1-1}
     After Task1: SNLI-VE & 59.0 & 60.5 & 59.8 & 60.6 & 26.8 & 27.3 & 23.3 & 26.5 & 52.7 & 55.2\\
     After Task1: VQAv2 & 58.4 & 59.6 & 58.6 & 60.2 & 26.4 & 26.7 & 22.5 & 23.6 & 50.8 & 51.8\\
     After Task1: NLVR2 & 56.2 & 56.5 & 55.8 & 58.2 & 26.0 & 27.1 & 21.5 & 25.4 & 52.8 & 54.1\\
     After Task1: VCR & 49.9 & 51.1 & 51.3 & 51.5 & 26.6 & 26.7 & 21.9 & 24.3 & 49.3 & 51.9\\
     \midrule
     \emph{CL: Sequential Fine-Tuning} \\
     \cmidrule(lr){1-1}
     After Task2: NLVR2 & 57.4 & 58.1 & 62.0 & 57.2 & 25.4 & 26.2 & 23.3 & 24.5 & 51.1 & 52.6\\
     After Task3: SNLI-VE & 52.3 & 56.8 & 60.2 & 63.3 & 26.1 & 26.4 & 22.2 & 23.8 & 51.1 & 53.2\\
     After Task4: VCR & 50.2 & 51.5 & 54.2 & 53.6 & 25.3 & 26.4 & 20.8 & 24.0 & 50.4 & 51.7\\
     \midrule
     \emph{CL: Experience Replay} \\
     \cmidrule(lr){1-1}
     After Task2: NLVR2 & 53.2 & 57.4 & 57.3 & 58.1 & 25.4 & 25.9 & 22.4 & 23.8 & 50.6 & 52.4\\
     After Task3: SNLI-VE & 57.1 & 58.6 & 61.8 & 61.9 & 25.7 & 26.0 & 22.8 & 23.6 & 51.2 & 51.8\\
     After Task4: VCR & 50.1 & 50.1 & 54.0 & 53.2 & 26.4 & 27.4 & 24.1 & 25.9 & 51.5 & 52.5\\  
     \midrule
     \emph{CL: EWC} \\
     \cmidrule(lr){1-1}
     After Task2: NLVR2 & 51.0 & 55.1 & 59.7 & 57.0 & 25.2 & 26.4 & 21.6 & 24.1 & 50.0 & 52.1\\
     After Task3: SNLI-VE & 53.9 & 54.5 & 57.1 & 57.5 & 25.7 & 26.1 & 20.6 & 22.4 & 50.0 & 52.8\\
     After Task4: VCR & 49.7 & 49.8 & 51.3 & 51.1 & 25.9 & 27.0 & 22.0 & 23.8 & 51.5 & 52.1\\   
     \midrule
     \emph{CL: Frozen Bottom-9} \\
     \cmidrule(lr){1-1}
     After Task1: VQAv2 & 57.4 & 57.9 & 56.5 & 58.9 & 25.8 & 26.9 & 22.5 & 27.3 & 50.3 & 54.4\\
     After Task2: NLVR2 & 53.9 & 55.7 & 56.2 & 58.7 & 26.6 & 27.2 & 24.8 & 27.0 & 51.3 & 54.0\\
     After Task3: SNLI-VE & 57.8 & 61.6 & 56.8 & 60.7 & 25.7 & 27.1 & 23.5 & 26.6 & 51.1 & 53.0\\
     After Task4: VCR & 54.2 & 55.7 & 56.5 & 58.2 & 26.1 & 27.2 & 24.3 & 27.4 & 51.5 & 53.6\\
     \midrule
     Random & 50.0 & 50.0 & 50.0 & 50.0 & 25.0 & 25.0 & 20.0 & 20.0 & 50.0 & 50.0 \\
     \bottomrule
    \end{tabular}
    
    \caption{Comparisons between different CL algorithms on language-only tasks. Each value is the average accuracy ($\%$) over 3 runs. We experiment with $\{16, 32\}$-shot per class on IMDB, SST-2. For HellaSwag, CommonsenseQA, and PIQA, we sub-sample $\{1024, 4096\}$ instances of training data.}
  \label{table:full-l}
\end{table}

\section{Experiments Using Another Vision-Language Model: UNITER}
We conduct CL experiments using UNITER~\cite{chen2019uniter} as the encoder. 
UNITER uses region features from a pre-trained Faster-RCNN as the visual input, in contrast to \vilt\ which directly operates on image patch tokens. 
We train UNITER on the same sequence of four upstream tasks (VQA $\rightarrow$ NLVR2 $\rightarrow$ SNLI-VE $\rightarrow$ VCR), using all of our CL algorithms except Adapters. 

In Table~\ref{tab:uniter}, we compare Upstream Knowledge Transfer between various CL algorithms, trained using the UNITER model. We see that UNITER, similar to \vilt, has negative transfer for post-VQA tasks, although UNITER typically has less negative transfer for the third task (SNLI-VE) than the second task (NLVR2).

%In Table~\ref{tab:uniter-forgetting}, we compare forgetting transfer between Sequential Fine-tuning and Experience Replay, when trained using the UNITER model. Similar to our findings with \vilt, we see that Experience Replay helps mitigate the forgetting issue, particularly on the VQA task.

\begin{figure}
    \centering
    \includegraphics[width=0.7\linewidth]{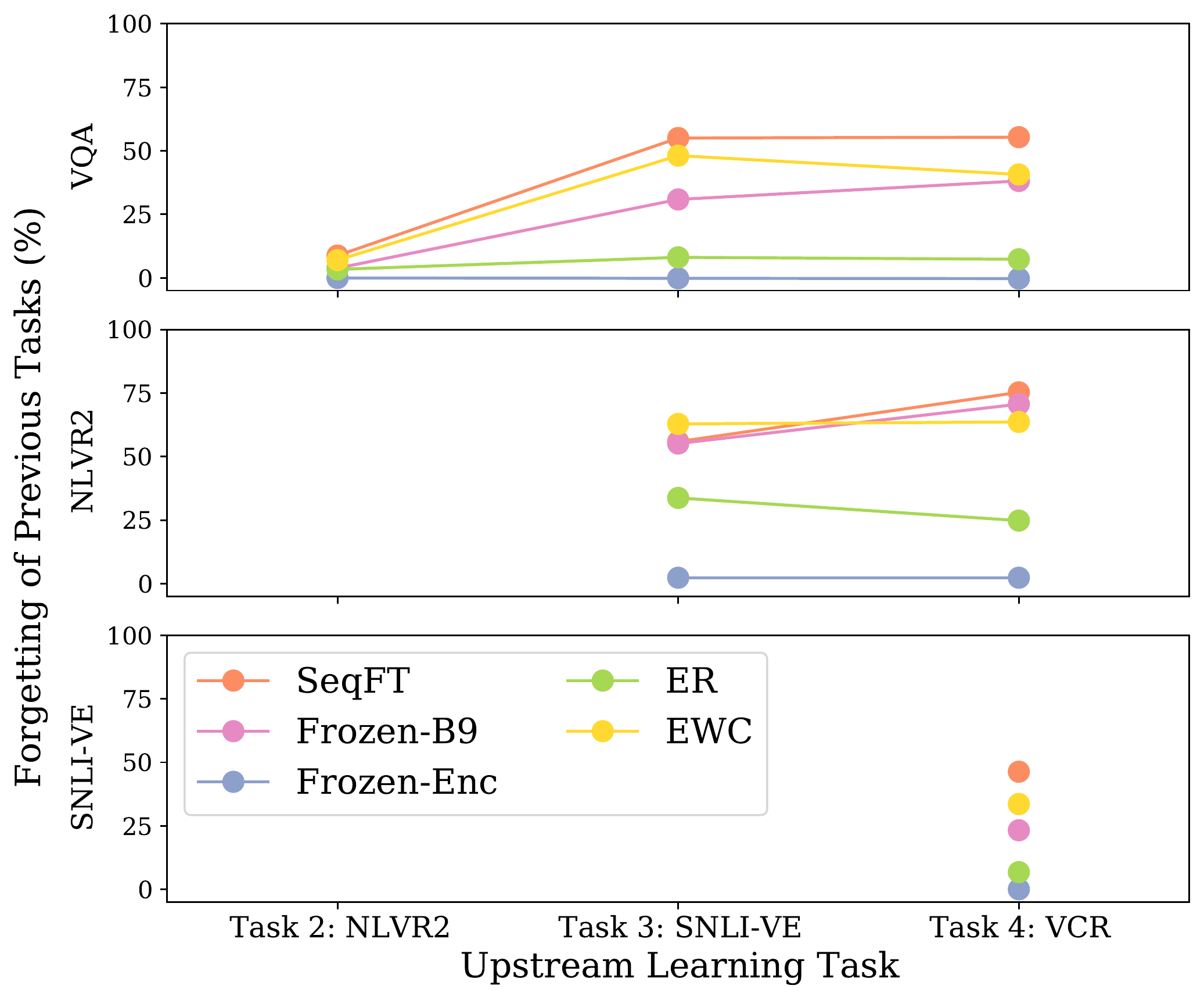}
    \caption{ Forgetting $\forgetting{j}{i}$ (\%) of the previous $i-1$ tasks when the UNITER model is trained using each algorithm. 
        Each subplot denotes model performance on one of the previous tasks. 
        Similar to our findings with \vilt, ER best preserves past task performance among all algorithms that fine-tune shared parameters.}
    \label{fig:uniter-forgetting}
\end{figure}

In Figure~\ref{fig:uniter-forgetting}, we plot Forgetting of previous tasks when the UNITER model is continually learned. We observe that Forgetting trends between algorithms are similar to our findings with \vilt: fine-tuning fewer shared parameters leads to less forgetting, while Experience Replay performs best among the algorithms that fine-tune all the shared parameters. 
In contrast to \vilt, we see that the VCR task does not impact the UNITER model's Forgetting as severely. This is likely due to the fact that UNITER (which utilizes region features) directly uses the ground-truth bounding boxes from the VCR task as part of the model, rather than drawing on the boxes onto the image as the image patch-based \vilt\ model does.

\begin{comment}

\begin{table}[t]
    \setlength{\aboverulesep}{2pt}
    \setlength{\belowrulesep}{2pt}
    \centering
    \begin{small}
    \begin{tabular}{llrrr}
    \toprule
    Pre-trained & \multirow{2}{*}{Alg $\algorithm$} &   \multicolumn{1}{c}{Task 1} & \multicolumn{1}{c}{Task 2} & \multicolumn{1}{c}{Task 3} \\
    VL Encoder& &  \multicolumn{1}{c}{VQAv2} & \multicolumn{1}{c}{NLVR2} & \multicolumn{1}{c}{SNLI-VE} \\
    \midrule
    \multirow{3}{*}{\vilt} & Direct FT &   [67.70]	& [73.07] & [76.31] \\
    \cmidrule(lr){2-5}
    & SeqFT  &   \transfer{0.13}{67.79}	& \transfer{-1.80}{72.66} & \transfer{-3.33}{74.89} \\
    & ER &   \transfer{0.26}{67.87} & \transfer{0.56}{73.20} & \transfer{-2.89}{75.08} \\
    \midrule
    \multirow{3}{*}{UNITER} & Direct FT &   [69.98]	& [75.82] & [78.09] \\
    \cmidrule(lr){2-5}
    & SeqFT  &   \transfer{0.05}{70.01}	& \transfer{-4.71}{74.61} & \transfer{-1.59}{77.38} \\
    & ER &   \transfer{-0.11}{69.90} & \transfer{-1.44}{75.45} & \transfer{-1.43}{77.45} \\
    \bottomrule
    \end{tabular}
    \end{small}
    \caption{
        Comparison of Upstream Knowledge Transfer $\knowledgetransfer{i}$ relative to direct fine-tuning on each task, along with task score $[S_{\algorithm}^i]$ (\%), between \vilt\ and UNITER for two CL algorithms.
        We see that UNITER does not have as much negative transfer on the SNLI-VE task as \vilt. 
    }
\label{tab:uniter}
\end{table}

\end{comment}

\begin{table}[t]
    \setlength{\aboverulesep}{0pt}
    \setlength{\belowrulesep}{0pt}
    \centering
    \begin{small}
    \begin{tabular}{lrrrr}
    \toprule
    \multirow{2}{*}{Alg $\algorithm$} &  \multicolumn{1}{c}{Task 1} & \multicolumn{1}{c}{Task 2} & \multicolumn{1}{c}{Task 3} & \multicolumn{1}{c}{Task 4} \\
    & \multicolumn{1}{c}{VQAv2} & \multicolumn{1}{c}{NLVR2} & \multicolumn{1}{c}{SNLI-VE} & \multicolumn{1}{c}{VCR} \\
    \midrule
    %\multirow{6}{*}{ViLT} & Direct FT &  [67.70]	& [73.07] & [76.31] & [61.31] \\
    %\cmidrule{2-6}
    %& SeqFT  &  \transfer{0.13}{67.79}	& \transfer{-1.80}{72.66} & \transfer{-3.33}{74.89} & \transfer{-5.09}{59.47} \\
    %& Frozen Enc &  \transfer{-14.10}{58.15} & \transfer{-40.78}{63.66} & \transfer{-15.98}{69.45} & \transfer{-53.47}{41.90} \\
    %& Frozen B9 &  \transfer{-0.58}{67.30} & \transfer{-0.58}{72.94} & \transfer{-3.31}{74.90} & \transfer{-15.49}{55.69} \\
    %& ER &  \transfer{0.26}{67.87} & \transfer{0.56}{73.20} & \transfer{-2.89}{75.08} & \transfer{-4.45}{59.70} \\
    %& EWC & \transfer{0.20}{67.84} & \transfer{-2.79}{72.39} & \transfer{-4.52}{74.38} & \transfer{-4.86}{59.55} \\
    %\midrule
    Direct FT &  [69.30]	& [75.25] & [78.09] & [69.97] \\
    \cmidrule{2-5}
    SeqFT  &  \transfer{0.06}{69.34}	& \transfer{-4.76}{74.05} & \transfer{-3.69}{76.44} & \transfer{-11.92}{64.61} \\
    Frozen Enc &  \transfer{-22.67}{53.59} & \transfer{-58.82}{60.40} & \transfer{-24.23}{67.24} & \transfer{-48.85}{48.01} \\
    Frozen B9 &  \transfer{-0.89}{69.91} & \transfer{-5.16}{73.95} & \transfer{-0.68}{77.79} & \transfer{-9.80}{65.57} \\
    ER &  \transfer{0.06}{69.34} & \transfer{-3.35}{74.41} & \transfer{-2.26}{77.07} & \transfer{-11.24}{64.92} \\
    EWC & \transfer{0.06}{69.34} & \transfer{-4.71}{74.06} & \transfer{-3.52}{76.51} & \transfer{-12.41}{64.39} \\
    \bottomrule
    \end{tabular}
    \end{small}
    \caption{
        Upstream Knowledge Transfer $\knowledgetransfer{i}$ relative to direct fine-tuning on each task, along with task score $[S_{\algorithm}^i]$ (\%), for different CL algorithms $\algorithm$ applied to UNITER.
        %No CL algorithms achieve notable positive Knowledge Transfer, while the majority in fact \emph{hurt} learning of new tasks.
    }
\label{tab:uniter}
\end{table}

\section{Hardware and Resources Used}
Our experiments were performed on an Exxact workstation containing four NVIDIA RTX 3090 GPUs.
Each upstream continual learning experiment was run on a single GPU.
While individual tasks took between 12 hours and 2 days to train, the entire 4-task continual learning typically took between 4 and 5 days.
For downstream experiments, each (upstream checkpoint, downstream task, sample size, random seed) run took between 30 minutes to 3 hours to train, depending on the tasks.

\end{document}